\documentclass[10pt,twocolumn]{article}

\usepackage[margin=2cm]{geometry}
\usepackage{amsmath}
\usepackage{amssymb}
\usepackage{graphicx}
\usepackage{booktabs}
\usepackage[numbers]{natbib}
\usepackage{hyperref}

\title{Triangulation-Free Bundle Adjustment with Graduated Non-Convexity
  for Camera Pose Refinement from Coarse Priors}
\author{Nikolaos Kyriazis\\
  \small nkyriazis@gmail.com ~$\cdot$~ ORCID 0000-0003-4124-520X}
\date{}

\begin{document}

\maketitle

\begin{abstract}
Mobile AR frameworks attach a metric pose prior to every casual phone
capture, and turning it into reconstruction-grade poses cheaply on CPU
is the step before novel-view synthesis. The least a refiner
owes an accurate prior is not to make it worse. The workhorse refiner
does. On 15 ScanNet++ iPhone room captures, COLMAP triangulation plus
prior-seeded bundle adjustment degrades an accurate ARKit prior in all
15, $0.55^{\circ}$ to $0.74^{\circ}$ by scene-mean.

The cause is the seeding. Structure is triangulated from the prior
before anything is optimized, so the prior's error is baked into the
structure the optimizer trusts. We remove the triangulation. Every
keypoint owns a scalar depth along its own back-projected ray and each
match contributes two symmetric cross-projection residuals, so structure
is re-expressed at every iterate. The same solve holds the room prior at
$0.57^{\circ}$ and never fails in 330 perturbed room runs, and at
object scale reaches
$0.265^{\circ}$/1.80\,mm from a $0.456^{\circ}$ prior in a median of
10\,s per scene on one CPU, against 2.5 GPU-hours for a learned refiner.

Because no structure is committed, the objective also admits graduated
non-convexity, which measures how deep the defect goes. Classical
refinement collapses past $1$--$2^{\circ}$ of prior error, barely beyond
a real ARKit prior, and no classical refinement arm survives
$32^{\circ}$. Ours
recovers 425 of 425 runs through $16^{\circ}$/80\,mm and 85\% at
$32^{\circ}$/160\,mm, and perturbed rooms through $32^{\circ}$.

Nominal object-scale accuracy is on par rather than better, on a
benchmark at its own noise floor, where classical bundle adjustment is a
strong baseline absent from the literature. One scene fails for every
solver already at zero perturbation. Re-mapping from position priors
matches us in the prior's frame but discards it, so it cannot exploit a
prior worth keeping or be warm-started.
\ifdefined\IJCVBUILD
\fi
\end{abstract}

\section{Introduction}

\begin{figure*}[t]
  \centering
  \includegraphics[width=\textwidth]{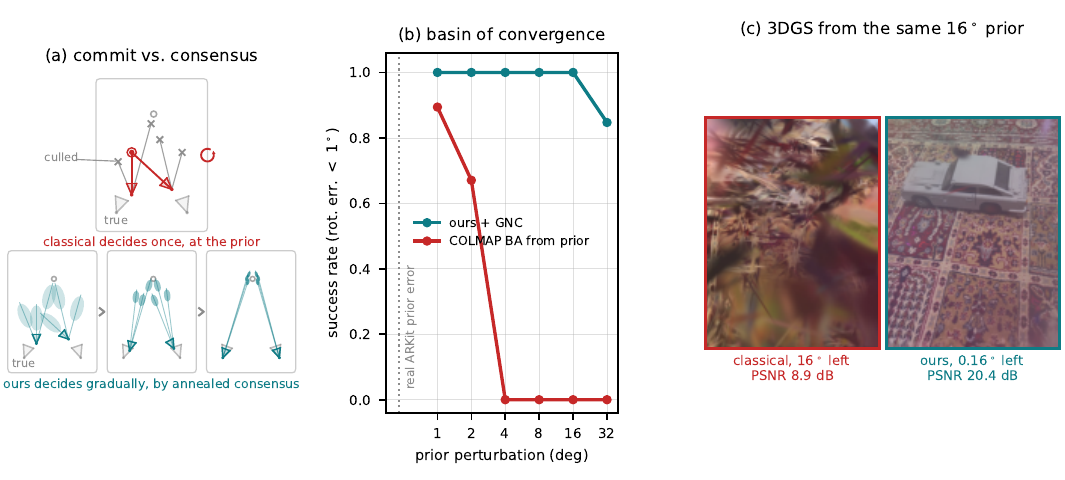}
  \caption{The claim at a glance. \textbf{(a)} Classical refinement
    decides matches once, at the prior. Triangulation stamps its
    verdict where the prior's rays cross and culls the matches that
    disagree, among them the ones pointing at the ghosted truth, so
    iterating returns the same wrong state. Ours never hard-commits.
    Every match keeps a soft vote along its own ray, and annealing
    the loss scale makes the consensus more pronounced, discarding
    nothing, until poses and associations settle on the truth
    together. \textbf{(b)} Success rate against prior
    perturbation on the 17 MobileBrick scenes whose matches admit the
    true solution, five seeds per scene (castle is the failure case,
    Section~\ref{sec:exp-failure}; protocol in
    Section~\ref{sec:exp-basin}). Classical prior-seeded BA collapses
    beyond $1$--$2^{\circ}$; ours with GNC recovers every run through
    $16^{\circ}$ and 85\% at $32^{\circ}$. The dotted line is the mean
    real ARKit prior error, $0.46^{\circ}$. \textbf{(c)} 3D Gaussian
    Splatting trained on each solver's result from the identical
    $16^{\circ}$ perturbed prior (aston, seed 0), rendered at
    the same held-out view.}
  \label{fig:teaser}
\end{figure*}
\label{sec:intro}

Mobile AR frameworks such as ARKit~\citep{apple_arkit} and
ARCore~\citep{google_arcore} attach a per-frame metric pose to any
casual video capture. Novel-view synthesis with
NeRF~\citep{mildenhall2020nerf} or 3D Gaussian
Splatting~\citep{kerbl20233dgs} needs poses better than those priors
supply, because small pose errors translate directly into blur and
ghosting, so a refinement stage sits between capture and reconstruction.
This paper is about doing that step cheaply, on CPU, from AR data, and
the first thing such a refiner owes an accurate prior is not to damage
it.

The workhorse refiner does damage it. On 15 ScanNet++ iPhone room
captures, COLMAP triangulation plus prior-seeded bundle
adjustment~\citep{schoenberger2016colmap} makes an already-accurate
ARKit prior worse in all 15 scenes, from a $0.55^{\circ}$ scene-mean to
$0.74^{\circ}$, at zero perturbation and with no adversarial setup
(Section~\ref{sec:exp-scannetpp}). The cause is the seeding rather than
the optimization. Structure must first be triangulated from the prior
poses, so the prior's error is baked into the structure that the
subsequent optimization then trusts.

The same defect has a sharper diagnostic. If triangulation commits the
prior's error, then refinement should fail not only subtly on a good
prior but completely on a coarse one, and it does. On
MobileBrick~\citep{li2023mobilebrick}, classical triangulation plus
bundle adjustment collapses at its documented configuration once the
prior errs by more than ${\sim}1$--$2^{\circ}$, and by
${\sim}8^{\circ}$ under any retuning of its gates and depth we tested
(Section~\ref{sec:exp-basin}); beyond its basin it returns the bad
prior essentially unchanged (Figure~\ref{fig:teaser}). ARKit's real
error (${\sim}0.46^{\circ}$) sits just inside that basin, which is
precisely why prior-seeded classical BA looks excellent on this
benchmark, and a prior a few times worse would already defeat it. Image
retrieval, GPS plus compass, and drifting visual-inertial odometry all
supply priors in exactly that range. At the other extreme, learned
refiners such as BARF~\citep{lin2021barf},
SPARF~\citep{truong2023sparf} and PoRF~\citep{bian2024porf} couple pose
refinement to training a neural scene representation and cost GPU-hours
per scene.

We propose a bundle adjustment with no triangulation to poison. Each
observed keypoint owns a single scalar depth along its own back-projected
ray, and each match induces two symmetric cross-projection residuals
between the implied 3D points and the observed pixels. This keeps the
familiar robust reprojection-error form of bundle adjustment while
removing track building and triangulation entirely. Structure is never
committed from the initial poses; it is re-expressed by every solver
iterate.

That one change answers the question the paper opened with. On the same
15 rooms the plain solve holds the accurate ARKit prior at
$0.57^{\circ}$ where the classical pipeline degrades it to
$0.74^{\circ}$, and no solve of ours failed anywhere in the 330-run
perturbed room campaign. At object scale it reaches
$0.265^{\circ}$/1.80\,mm from a $0.456^{\circ}$ prior at a median of
10\,s per scene on a single CPU with capped matches, or
$0.246^{\circ}$/1.61\,mm at 101\,s uncapped, against GPU-hours for the
learned refiners. Downstream, splatting on those poses closes half the
prior-to-ground-truth PSNR gap.

Because structure is never committed, the objective also admits
graduated non-convexity (GNC)~\citep{blake1987visual,yang2020gnc},
which anneals the robust-loss scale over warm-started stages and
smooths, then progressively sharpens, the cost landscape. This is what
turns the diagnosis into a measurement. The resulting basin of
convergence covers every feasible run through $16^{\circ}$/80\,mm of
prior error and 85\% of feasible runs at $32^{\circ}$/160\,mm, about
$35$--$70\times$ ARKit's real error, more than an order of magnitude
beyond where triangulation-based BA as documented fails and a factor of
two beyond its strongest modernized refinement arm, at nominal accuracy
indistinguishable from our plain solve. At room scale, once basin
traversal is staged before self-calibration, the annealed variant
recovers perturbed priors through $32^{\circ}$ at a median
indistinguishable from nominal, at a nominal scene-mean price
($0.69^{\circ}$, concentrated in one apartment;
Section~\ref{sec:exp-scannetpp}).

Our contributions are stated relative to well-established prior art
(Section~\ref{sec:related}).
\begin{itemize}
  \item Refinement that does not damage an accurate prior, which is the
    property the motivating application needs first and the one the
    standard pipeline lacks. On 15 ScanNet++ iPhone rooms classical
    triangulation plus bundle adjustment degrades an ARKit-grade prior
    in all 15 at zero perturbation, while our solve holds it and
    completes the 330-run perturbed room campaign without a single
    solver failure. At object scale it refines a $0.456^{\circ}$ prior to
    $0.265^{\circ}$/1.80\,mm at a median of 10\,s per scene on one CPU
    with capped matches, or $0.246^{\circ}$/1.61\,mm at 101\,s uncapped,
    and closes half the prior-to-ground-truth PSNR gap downstream in
    splatting.
  \item A prior-anchored bundle adjustment that commits no structure at
    initialization, which is why the above holds. Every observation
    carries its own scalar depth along
    its own back-projected ray, with no landmark variables, symmetric
    pairwise cross-projection residuals and optional self-calibration of
    shared intrinsics and distortion, so structure is re-expressed by
    every solver iterate instead of fixed once from the prior. That
    continuous re-expression is the mechanism the basin results turn on.
    The form is also cheap. Each eliminated Schur block is $1\times 1$,
    each residual couples exactly two cameras, and a shared evaluation
    cache reduces the per-residual cost to a single projection, which
    makes it the fastest CPU variant we evaluate (median 10\,s per scene
    with capped matches, single CPU). Per-observation depths are one way
    to obtain re-expression and not the only one; anchored tracks,
    re-expressed the same way, match them level for level under GNC
    (Table~\ref{tab:basin-controls}). The structural core has dense and
    learned relatives (Section~\ref{sec:related}); what is new, to our
    knowledge, is the combination as a standalone classical refiner over
    detected sparse matches.
  \item Robustness to coarse priors via GNC. Annealing the robust-loss
    scale over warm-started stages recovers every feasible run from
    prior perturbations through $16^{\circ}$/80\,mm, about
    $35\times$ the real error of an ARKit prior (425/425 runs across 17
    scenes and 5 noise seeds; the 18th fails already at zero
    perturbation), recovers 85\% of feasible runs at
    $32^{\circ}$, and
    eliminates the plain solve's seed-dependent edge failures, with
    nominal accuracy unchanged. The same annealing has no handle in
    triangulation-based BA as deployed, whose structure is committed by
    threshold-filtered triangulation before any loss is evaluated and
    whose basin we measure at
    ${\sim}1$--$2^{\circ}$ under identical perturbations, at
    ${\sim}8^{\circ}$ under gate-and-depth tuning, and at $16^{\circ}$
    in shape,
    no longer deterministically, when we port the annealing itself into
    a current build through robust losses and staged re-triangulation.
  \item Refinement that keeps the prior's metric frame, and a
    measurement of who else does. Sim(3) alignment with scale, the
    standard protocol, forgives exactly the gauge and scale that a
    metric AR prior exists to supply, so we also score every arm with
    no alignment at all. Classical refinement is gauge-free and
    therefore never anchors, inheriting nearly the whole prior error in
    the absolute frame and, at an accurate prior, leaving the poses
    further from truth than not refining. Prior-position mapping does
    anchor, but its Sim(3) score is identical to three decimals across
    a $32\times$ change in prior quality, which settles that it
    reconstructs and registers rather than refines. Our solve matches
    its absolute error to within ten percent at $8^{\circ}$,
    $16^{\circ}$ and $32^{\circ}$, so refinement forfeits nothing in
    the prior's own frame relative to discarding that prior, and the
    choice between them rests on the functional and reliability
    differences rather than on accuracy
    (Section~\ref{sec:exp-frame}).
  \item A strong classical baseline, and the reason it is brittle.
    Prior-seeded classical BA (COLMAP triangulation plus BA from ARKit
    priors) reaches 0.218$^{\circ}$/1.50\,mm on MobileBrick at a median of
    53\,s per scene on CPU, on par in rotation with PoRF
    (0.22$^{\circ}$/1.23\,mm at 2.5 GPU-hours per scene). To our knowledge
    this baseline is unreported on this benchmark. We both establish it
    and delimit it, showing its accuracy is conditional on a prior already
    inside its narrow basin.
  \item An evaluation that reports its own weak points. Per-scene results
    on all 18 MobileBrick evaluation scenes, headline means reported both
    with and without the prior trust region (whose thresholds are derived
    from MobileBrick training scenes, not the test set), a
    basin-of-convergence study of all three solvers over 18 scenes and
    five noise seeds, a room-scale campaign on 15 ScanNet++ scenes (735
    runs; classical refinement degrades the prior in all 15 at zero
    perturbation), a mechanism study (Section~\ref{sec:exp-mechanism}; measured 1-D cost landscapes
    and an oracle-structure control that pins the classical collapse on
    triangulation from the prior), a downstream splatting control, a
    real-drift characterization on LaMAR, and a failure analysis of the
    one scene where the sparse matches themselves prefer a wrong,
    coherent solution.
\end{itemize}

We will release the code.

\section{Related work}
\label{sec:related}

\paragraph{Point parameterizations in bundle adjustment.}
Classical bundle adjustment parameterizes structure as 3D points shared by
all observations of a track. Civera et
al.~\citep{civera2008inverse} introduced the inverse-depth parameterization
for monocular SLAM, where a landmark is a depth along a ray anchored at
the first observing camera, which linearizes measurement equations for
low-parallax features. ParallaxBA~\citep{zhao2015parallaxba} replaces depth
with a parallax angle between two anchor views, improving conditioning of
the normal equations when features are distant or observed with small
baselines. Both retain one structure parameter set per landmark. Our
parameterization goes one step further, with one depth per observation,
anchored at that observation's own camera, and no landmark variable at
all. Consistency between the two sides of a match is enforced by symmetric
cross-projection residuals rather than by a shared variable.
Per-observation ray scales do appear elsewhere with different roles.
Projective factorization~\citep{sturm1996factorization} carries one
projective depth per observation, but as a scale on a tracked measurement
matrix under a global rank-four constraint.
GLOMAP's~\citep{pan2024glomap} global-positioning step assigns each
observation its own scale along its ray inside a convex surrogate and then
reverts to landmark BA for refinement, and pose-only
formulations~\citep{cai2021poseonly} write a depth per observation only to
eliminate it analytically. In our objective the per-observation depths
remain the optimized structure proxy throughout.

\paragraph{Structureless bundle adjustment.}
A complementary line eliminates structure analytically. Incremental Light
Bundle Adjustment~\citep{indelman2012ilba} replaces reprojection errors with
three-view constraints so that only camera poses are optimized; reduced
epipolar costs~\citep{rodriguez2011gea} and pose-only ``epipolar
adjustment''~\citep{fastmap2025} pursue the same elimination at scale, and
the filtering literature marginalizes landmarks per update (MSCKF's
null-space projection~\citep{mourikis2007msckf}, smart
factors~\citep{carlone2014smart}). Cefalu et
al.~\citep{cefalu2016structureless} formulate structureless BA with
self-calibration through accumulated epipolar and scale-consistency
constraints. These methods trade the reprojection-error form, and with it
the mature robust-loss machinery of BA, for algebraic constraints among
cameras. We keep a minimal 1-DoF structure proxy per observation, which
preserves the standard reprojection residual and its pixel-space robust
losses, and also lets the same objective self-calibrate intrinsics
and distortion. Concurrently with this work, Marginalized Bundle
Adjustment~\citep{zhu2026mba} builds a triangulation-free objective around
dense monocular-depth estimates; it targets pose estimation from learned
depth rather than prior refinement, and uses no loss annealing.

\paragraph{Initialization-free bundle adjustment.}
A further line widens or removes the basin of convergence by changing the
objective itself. VarPro-based projective BA converges from arbitrary
initializations~\citep{hong2016varpro}, pOSE interpolates object-space and
affine surrogates with a wide basin~\citep{hong2018pose}, and PoVar scales
the approach to large problems~\citep{weber2024povar}, with recent
extensions reaching near-metric calibrated
solutions~\citep{olsson2026calibba}. ProBA models landmarks as 3D Gaussians
to enlarge the basin from cold starts~\citep{chui2025proba}, and XM lifts
each observation to its own scaled depth inside a convex
relaxation~\citep{han2025rome}, structurally close to our
parameterization. These methods reconstruct from scratch, and they buy
their basin by giving up exactly what our setting supplies and needs, a
metric prior that anchors gauge and scale and a robust pixel-space
residual against outlier-contaminated exhaustive matches (pOSE-family
objectives are least-squares on curated tracks). We keep both, and we
measure the basin against prior error rather than against random starts.

\paragraph{Known-rotation and $L_\infty$ formulations.}
Once the rotations are fixed, camera positions and structure become
quasi-convex in the remaining variables under the $L_\infty$ norm of
reprojection error, and a bisection over second-order cone feasibility
tests returns the global optimum without an
initialization~\citep{hartley2004linf,kahl2005linf,kahl2008linf}, with
the known-rotation motion problem treated
directly by~\citet{sim2006linf}. There is no basin to speak of in that
family, and our regime is closer to its assumptions than it may appear.
The match-graph bootstrap of Section~\ref{sec:exp-basin} delivers global
rotations at $0.34^{\circ}$ on aston from correspondences alone, using
classical rotation
averaging~\citep{hartley2013rotavg,chatterjee2013rotavg,dellaert2020shonan}.
Three properties keep it from replacing the setting studied here. It
needs rotations that are already good, whereas the problem we pose is a
prior whose rotations are wrong by degrees and whose rotational
correction is the output, and rotations enter a known-rotation program as
constants it never refines. The $L_\infty$ cost is decided by the single
worst residual, so it is outlier-sensitive by construction on the
exhaustive, uncurated matches we solve over, and the standard remedy of
iteratively removing the maximal-residual measurements is a discrete
commitment of the kind this paper is trying to avoid. We state plainly
that for the sub-$8^{\circ}$ regime, with rotation averaging in front to
supply the rotations, a known-rotation $L_\infty$ pipeline is a
legitimate alternative that we did not run.

\paragraph{Robust optimization and graduated non-convexity.}
Graduated non-convexity, the optimization of a sequence of objectives from
smooth to sharp, goes back to Blake and Zisserman~\citep{blake1987visual}
and was revived for spatial perception by Yang et
al.~\citep{yang2020gnc}, who anneal a surrogate robust loss to reject
outliers in registration and pose-graph problems without an initial guess.
In bundle adjustment, robust losses are standard but their scale is
conventionally fixed. Lifting~\citep{zach2014robust}, graduated smoothing
of the kernel~\citep{zach2018descending,le2020graduated} and adaptive
kernels~\citep{chebrolu2021adaptive,jung2024agnc} adjust the effective
robustness during the solve to escape poor local minima or fit the
residual distribution, in landmark-based BA and targeting outlier
robustness. Outside BA, annealed and even adaptively scheduled loss scales
are established for averaging and registration; Sidhartha et al.\ adapt
the schedule by tracking curvature~\citep{sidhartha2023annealing,
sidhartha2024averaging}, convergence theory exists for the underlying
IRLS iterations~\citep{peng2023irls}, and the certifiable-perception line
pursues global optimality outright~\citep{yang2022certifiable}.
BARF~\citep{lin2021barf} applies the same coarse-to-fine principle in the
photometric domain by annealing positional encodings. We apply loss-scale
annealing to a structureless sparse objective and measure its effect on
the basin of convergence directly. The combination matters because
triangulation-based BA cannot benefit from it; its structure is committed
from the initial poses before any loss is evaluated
(Section~\ref{sec:exp-basin}).

\paragraph{Joint pose--radiance-field optimization.}
A recent family refines poses jointly with a neural scene representation.
BARF~\citep{lin2021barf} anneals positional encodings to widen the basin of
convergence of NeRF-based pose registration;
SCNeRF~\citep{jeong2021scnerf} adds a projected-ray-distance correspondence
loss, a loss between rays without shared 3D points that is conceptually
close to our residual but evaluated inside a NeRF training loop.
SPARF~\citep{truong2023sparf} exploits dense correspondences and depth
consistency for sparse-view settings. The same joint refinement has since
moved to 3D Gaussian Splatting, with preconditioned camera
optimization~\citep{park2023camp}, pose-free splatting from
video~\citep{fu2024cf3dgs}, and sparse-view pipelines seeded by learned
pointmaps~\citep{fan2024instantsplat}. Closest to us in problem statement,
PoRF~\citep{bian2024porf} trains a per-scene MLP pose-residual field starting
from ARKit (or COLMAP) priors and reports pose accuracy on
MobileBrick~\citep{li2023mobilebrick}; training takes hours on a GPU. Our
experiments address the same prior-refinement problem with sparse
second-order solves that run in seconds to minutes on CPU, and we adopt
PoRF's MobileBrick protocol for comparison.

\paragraph{Per-observation depths in dense and learned systems.}
The structural core of our parameterization, a depth per observation with
consistency enforced only through residuals, has dense and learned
relatives. DROID-SLAM~\citep{teed2021droid} optimizes per-pixel per-frame
inverse depths with Gauss--Newton and Schur elimination over the depth
block, against dense correspondences re-predicted by a network each
iteration, with learned confidence weights in place of a robust kernel;
MegaSaM~\citep{li2025megasam} and ViPE~\citep{huang2025vipe} inherit the
same solver structure and add monocular depth priors.
RidgeSfM~\citep{graham2020ridgesfm} also skips tracks and optimizes poses
against sparse pairwise matches, but confines depth freedom to a
low-dimensional learned basis per frame.
DUSt3R~\citep{wang2024dust3r} aligns free per-pixel 3D pointmaps (three
DoF per pixel, 3D residuals) by first-order descent.
MASt3R-SfM~\citep{duisterhof2025mast3rsfm}, the closest existing objective
to ours, minimizes a symmetric bidirectional reprojection loss over sparse
pixel matches in which each pixel is lifted by its own depth; its depths
are coupled to a coarse anchor grid, its robust loss is fixed
($\lVert x\rVert^{0.5}$), and it optimizes with Adam from learned-pointmap
initializations rather than with a second-order solver from a metric
prior. None of these anneals the loss or studies robustness to prior
error. When no prior exists at all, poses can be recovered from scratch,
incrementally with COLMAP~\citep{schoenberger2016colmap}, with
detector-free matching throughout~\citep{he2024dfsfm}, globally with
GLOMAP~\citep{pan2024glomap}, by incrementally learning a
relocalizer~\citep{brachmann2024ace0}, or feed-forward with
VGGT~\citep{wang2025vggt} and its
successors~\citep{wang2026pi3,pan2026gluemap}. That regime is complementary to ours, and
feed-forward poses themselves need refinement of exactly the kind studied
here. Classical refinement also exists downstream of reconstruction.
Pixel-perfect SfM~\citep{lindenberger2021pixsfm} polishes the poses and
points of an initial SfM model by featuremetric alignment against dense
deep features, with no per-scene training; it refines an existing
reconstruction and keeps its triangulated structure, whereas our problem
starts from a coarse metric prior with no structure to keep, and its
concern is accuracy at the optimum rather than the width of the basin
that reaches it.

\section{Method}
\label{sec:method}

\subsection{Problem setup}
\label{sec:setup}

The input is a casually captured phone video with per-frame camera pose
priors from a mobile AR framework
(ARCore/ARKit~\citep{google_arcore,apple_arkit}), rough shared intrinsics,
and sparse feature matches between frames. The output is a set of refined
camera poses, optionally with refined intrinsics and distortion, accurate
enough for novel-view synthesis. It is obtained in a single sparse
bundle-adjustment solve with no triangulation, no track building and no
incremental mapping.

Camera $i$ has a camera-to-world pose with rotation $R_i$ and position
(ray origin) $O_i$. All cameras share pinhole intrinsics, focal lengths
$f$ and principal point $c$, and an OpenCV-style distortion vector
$d = (k_1, k_2, p_1, p_2)$. Feature matching produces a set of $M$ matches;
a match associates observation $(i,k)$ (camera $i$, feature $k$, distorted
pixel $u_{ik}$) with observation $(j,l)$.

\subsection{Ray-time parameterization}
\label{sec:param}

For every observed keypoint $(i,k)$ the observation is first undistorted,
$\hat u_{ik} = \mathrm{undistort}(u_{ik}; f, c, d)$, by inverting the
OpenCV $k_1,k_2,p_1,p_2$ model with Newton iterations in normalized camera
coordinates (up to 5 iterations, differentiable). The undistorted pixel
back-projects to a unit ray in world space with origin $O_i$ and direction
$D_{ik} = R_i \, \mathrm{ray}(\hat u_{ik}; f, c)$. The 3D point implied by
the observation is
\begin{equation}
  X_{ik} = O_i + t_{ik} D_{ik},
  \label{eq:raypoint}
\end{equation}
where the scalar ray time $t_{ik} > 0$ is a free parameter owned by this
observation and shared with no track. It is initialized to $t_{ik} = 1$ in
metric meters, a sane default given the metric priors.

There is no landmark variable. A match $(i,k) \sim (j,l)$ produces two
symmetric cross-projection residuals
\begin{align}
  r_{ij} &= \pi_j(X_{ik}) - \hat u_{jl}, \label{eq:rij}\\
  r_{ji} &= \pi_i(X_{jl}) - \hat u_{ik}, \label{eq:rji}
\end{align}
where $\pi_j$ projects a world point into camera $j$ with the shared
pinhole intrinsics. The comparison is against the undistorted observed
pixel, so distortion enters the problem only through the undistortion of
observations. Each residual stays a plain pinhole projection instead of
composing the forward distortion model per residual.

\subsection{Objective}
\label{sec:objective}

With $M$ matches and $N$ cameras, the total energy is
\begin{multline}
  E = \frac{1}{4M} \sum_{\text{matches}}
      \Big[ \rho\big(\lVert r_{ij}\rVert^2\big)
          + \rho\big(\lVert r_{ji}\rVert^2\big) \Big] \\
    + \frac{\lambda_p}{N} \sum_{i} \big\lVert O_i - O_i^{\text{prior}} \big\rVert^2 \\
    + \frac{\lambda_R}{N} \sum_{i} \big\lVert \operatorname{Log}\big( (R_i^{\text{prior}})^{-1} R_i \big) \big\rVert^2 .
  \label{eq:energy}
\end{multline}
Here $\rho(s) = a \arctan(s/a)$ is the arctan robust loss with scale
$a = 100$ in squared-pixel units by default. The loss acts on the squared residual
norm, so saturation sets in around 10\,px of reprojection error. Because
$\rho$ saturates, outlier matches degrade gracefully; there is no shared
landmark for them to poison. The position-prior term anchors the gauge (a
global similarity) to the metric AR frame and prevents trajectory
collapse; $\lambda_p = 10^{-2}$ by default, normalized by the camera count
$N$, while the match term is normalized by the residual count $4M$. The
third term softly anchors orientations; $\operatorname{Log}$ is the
small-angle rotation vector of $(R_i^{\text{prior}})^{-1} R_i$
(shorter-arc sign convention), and $\lambda_R = 10^3$ by default
(Section~\ref{sec:variants} motivates the value). Rotations are optimized
on the quaternion manifold; the
ray times are bounded below at $10^{-9}$ and the focal length is bounded
positive. Equation~\eqref{eq:energy} is the complete energy optimized in
all headline results.

\subsection{Structure of the normal equations}
\label{sec:structure}

The parameter blocks are the $2+2+4$ shared intrinsics (focal, principal
point, distortion), $7$ per camera (quaternion and position), and one per
observation (its ray time). The Schur complement eliminates the $t$
variables exactly as classical BA eliminates points, except that each
eliminated block is $1\times 1$ (vs.\ $3\times 3$ for a 3D landmark).
Each residual couples exactly two cameras. A ray time couples its own
camera to the cameras its observation is matched into (exactly one, for
the many observations that appear in a single pair), so its elimination
produces fill-in among those cameras exactly as a landmark observed by the
same set would. The reduced camera system is then solved by direct sparse
Cholesky factorization.

\subsection{Graduated non-convexity}
\label{sec:gnc}

Because no structure is triangulated from the initial poses, the entire
cost landscape is re-evaluated at every iterate, which makes loss-scale
annealing~\citep{blake1987visual,yang2020gnc} directly applicable. (We
use the GNC name for familiarity; no stage of our schedule is convex, so
the schedule is graduated kernel smoothing in the descending-robust-kernel
lineage~\citep{zach2018descending,le2020graduated} rather than a
Blake--Zisserman construction with a solvable first stage.) We
solve a short sequence of problems with the arctan scale annealed
$a = 10^4 \rightarrow 10^3 \rightarrow 10^2$, each stage warm-started from
the previous stage's solution. At $a = 10^4$ the loss is near-quadratic
over a ${\sim}100$\,px range and the objective is smooth enough to pull
the solution from gross initialization error. The final stage is exactly
the nominal objective, so the fixed point of the schedule coincides with
the plain solve; empirically, nominal accuracy moves by at most
$0.005^{\circ}$/0.02\,mm per scene (matched cap-100 runs on the six
development scenes) at about twice the solve cost
(Section~\ref{sec:exp-runtime}).
The kernel family is a choice of availability rather than of theory, and
we say so. Our $\rho$ is Ceres's \texttt{ArctanLoss}, which the solver
ships and whose scale the schedule anneals directly, whereas the GNC
line~\citep{yang2020gnc} surrogates Geman--McClure or truncated least
squares and drives the surrogate with a principled $\mu$ update. We did
not implement either surrogate under that update, so we claim no
equivalence between the two constructions. What our own control does
show is that the load-bearing choice is the starting scale and not the
kernel family; two, three and five stages are indistinguishable, while
starting a decade lower already loses runs at $8^{\circ}$/40\,mm
(Section~\ref{sec:exp-basin}). The stages are fixed; adaptive
schedules~\citep{sidhartha2023annealing,jung2024agnc} are an orthogonal
refinement, and the fixed schedule already recovers every feasible run
through $16^{\circ}$/80\,mm in our experiments.
Triangulation-based BA as deployed has no analogous handle. Its structure
is committed by threshold-filtered triangulation before the first residual
is evaluated, and with a coarse
prior that structure is already wrong (Section~\ref{sec:exp-basin}).
Permissive, annealed re-triangulation does partially emulate the
schedule inside a classical pipeline; staged gate loosening with
robust losses on a current COLMAP build recovers most runs through
$16^{\circ}$ but no classical refinement arm survives $32^{\circ}$
(Section~\ref{sec:exp-basin}). Re-expressing structure per stage is
the operative mechanism, and the structureless objective is where
re-expression is continuous, cheapest and complete. GNC
widens the basin toward whatever solution the matches prefer, which is the
desired behavior except when the matches themselves are degenerate, as in
the failure case of Section~\ref{sec:exp-failure}; the trust-region
guardrail below applies unchanged.

\subsection{Variants and guardrails}
\label{sec:variants}

\paragraph{Anchored-track variant (ablation baseline).}
To isolate the effect of the per-observation parameterization we also
implement the rigid, Civera-style~\citep{civera2008inverse} anchored-depth
variant of the same objective. Union-find tracks are built over the match
graph, each track owns a single ray time along its anchor observation's
ray, and residuals project the anchor point into every other camera
observing the track. Solver machinery (loss, priors, caching) is identical
and the eliminated blocks remain $1\times 1$, though a track time couples
all cameras observing the track; only the structure variables change.

\paragraph{Orientation prior (default $\lambda_R = 10^3$).}
The nominally principled setting is $\lambda_R = 1/\sigma_R^2$ with
$\sigma_R$ the AR framework's rotation error in radians, approximately
$0.5^{\circ}$ for ARKit as measured on
MobileBrick~\citep{bian2024porf,li2023mobilebrick}, giving
$\lambda_R \approx 10^4$. Empirically that over-pins. Rotations lock to
the prior, the position field warps to compensate, and Sim(3) alignment
converts the warp back into apparent rotation error (rotation degrades
0.293$^{\circ} \rightarrow$ 0.322$^{\circ}$ on our development subset
while translation improves). One step softer, $\lambda_R = 10^3$
($\sigma_R \approx 1.8^{\circ}$), improves both axes over no prior at all
on the development subset (on the evaluation set translation improves and
rotation is flat, Table~\ref{tab:ablation}) and is the default in all
headline runs; $\lambda_R = 3\times 10^3$
already re-enters the trade-off on scenes with poor priors. The
development subset is six evaluation scenes (aston, audi, bridge, camera,
colosseum, london\_bus), so this hyperparameter was chosen with sight of
part of the test set; Table~\ref{tab:ablation} bounds what that sight
buys, since the untuned $\lambda_R{=}0$ variant differs from the headline
by $0.001^{\circ}$/0.05\,mm. The same term at diagnostic strengths
appears in the failure analysis of Section~\ref{sec:exp-failure}.

\paragraph{Prior trust region (system guardrail).}
If the refined solution leaves the prior's plausibility envelope, the
matches are deemed unreliable and the prior is returned unchanged. The
envelope is derived from data disjoint from the evaluation set. On 9
MobileBrick training scenes the solve never moves more than
$1.75^{\circ}$/10.8\,mm from the prior, and we set the threshold at twice
that envelope, $3.5^{\circ}$/22\,mm. (The recorded evaluation runs used
$3^{\circ}$/15\,mm; no scene falls between the two thresholds, so the
scoring is identical under either.) On the 18 evaluation scenes exactly
one scene trips it, castle, whose solve moves 65\,mm from the prior
while the orientation prior holds its rotations within $0.7^{\circ}$,
so the twist expresses itself through the positions
(Section~\ref{sec:exp-failure}); every other test scene stays inside the
training envelope itself, at most $0.9^{\circ}$/10.5\,mm. Because the guardrail is load-bearing for
the headline mean, we report the means both with and without it
(Table~\ref{tab:mobilebrick}). The envelope width is by construction
a bet on the prior's quality, since any correction larger than the
envelope must trip it; how to place that bet when the quality is
unknown is resolved by the pre-solve and posterior checks the
conclusion assembles into a deployment recipe.

\subsection{Mapping onto Ceres}
\label{sec:solver}

The solver builds on Ceres Solver~\citep{agarwal2023ceres}. The robust
loss $\rho$ is \texttt{ArctanLoss}, rotations use the
\texttt{EigenQuaternionManifold}, the ray-time and focal-length bounds
are box constraints, and the elimination group of
Section~\ref{sec:structure} is recovered by Ceres' automatic ordering,
which picks out precisely the ray times. The reduced camera system is
factorized via SuiteSparse, multithreaded. The annealed stages of
Section~\ref{sec:gnc} run on one problem object; matches, caches and
sparsity are built once, and each stage warm-starts from the previous
stage's solution.

\paragraph{Shared evaluation cache.}
Each observation is shared between residuals. Its ray point is projected
in one residual of its match while its undistorted pixel is the target of
the other, and an observation matched into several frames recurs across
matches. The expensive part of both uses, the Newton undistortion and the
unprojection to a ray, does not depend on which residual asks for it, so
evaluating it inside each residual would repeat the costliest work of an
evaluation. A shared evaluation cache (Ceres \texttt{EvaluationCallback})
instead computes each
observation's undistorted pixel and ray point once per evaluation point,
in practice once per iteration, using forward-mode automatic
differentiation (jets) over the 16-dimensional local parameter set of that
observation (intrinsics 8, pose 7, ray time 1). Each residual then
performs ``gradient surgery'' and copies the cached jet derivatives into
the correct slots of its own 23-dimensional residual jet (intrinsics 8,
source camera 7, ray time 1, target camera 7). The marginal cost per
residual is therefore a single projection rather than a full
undistort-and-unproject chain; the work is done once per observation, not
once per residual. The cache is TBB-parallel across observations.

\paragraph{Line search.}
One profiling note is worth recording. The positivity
bounds on focal length and ray times make the problem box-constrained,
which by default triggers Ceres' projected line search and costs one extra
full Jacobian evaluation per iteration. The bounds are enforced by
projection in the parameter update regardless, so we disable the line
search. This leaves the solution bitwise identical on 12 of the 18
evaluation scenes, with third-decimal differences elsewhere, and cuts the
mean solve time by a third.

\section{Experiments}
\label{sec:experiments}

\subsection{Setup}
\label{sec:exp-setup}

MobileBrick~\citep{li2023mobilebrick} provides iPhone-captured object
videos with per-frame ARKit poses and ground-truth geometry from known LEGO
CAD models; its 18 curated evaluation sequences are the protocol used by
PoRF~\citep{bian2024porf} for pose refinement from ARKit priors. We follow
that protocol. All methods start from the same ARKit priors, and errors are
measured against the dataset's refined ground-truth poses (Sim(3)
alignment, per-frame ATE). We validated our evaluation code by reproducing
PoRF's published ARKit row (0.456$^{\circ}$/1.904\,mm vs.\ their
0.46$^{\circ}$/1.90\,mm) before scoring our own runs. Correspondences are
COLMAP SIFT features (affine shape estimation, domain-size pooling, peak
threshold $10^{-2}$, first octave $-2$) with exhaustive guided matching,
where the guidance is the two-view epipolar geometry COLMAP estimates
from the putative matches themselves during geometric verification, not
any pose. The match databases are built once per scene from the images
alone, before any prior or perturbation enters the pipeline, and every
variant below, including the classical control, consumes the same
databases. All CPU timings in this paper are on one AMD Ryzen~9 9950X
(16 cores, 32 threads); the GPU baselines we run ourselves (MASt3R-SfM,
VGGT) use one NVIDIA RTX~5090. Headline runs of our solver optimize camera poses only, with
intrinsics and distortion held fixed; self-calibration is exercised only
in the diagnostics of Section~\ref{sec:exp-failure}.

\paragraph{Classical control (prior-seeded COLMAP BA).}
As a control we run classical bundle adjustment from the same priors on the
same feature databases, using COLMAP~3.7's \texttt{point\_triangulator} and
\texttt{bundle\_adjuster}~\citep{schoenberger2016colmap} seeded with the
ARKit poses, alternating two rounds of triangulation and bundle adjustment
with intrinsics held fixed. No component of our method is involved.

\subsection{Pose accuracy on MobileBrick}
\label{sec:exp-mobilebrick}

Table~\ref{tab:mobilebrick} reports mean rotation and translation errors
over the 18 evaluation scenes; per-scene results are in the appendix
(Table~\ref{tab:perscene}). Baseline numbers are the published values from
PoRF (Table~3 of~\citep{bian2024porf}).

We begin at the nominal prior, where the question is not whether a
refiner recovers but whether it helps at all. All prior-seeded refiners
land in one band.
Prior-seeded classical BA reaches 0.218$^{\circ}$/1.50\,mm, on par in
rotation with PoRF, at a median of 53\,s per scene on CPU; to our
knowledge this baseline is absent from the MobileBrick pose-refinement
literature. Our per-observation solver reaches 0.246$^{\circ}$/1.61\,mm
with the prior trust region active (Section~\ref{sec:variants}); on one
scene (castle) the solve falls back to the ARKit prior, which is scored as
our result there (Section~\ref{sec:exp-failure}). Without the fallback,
castle's raw twisted solution enters the mean, which becomes
0.488$^{\circ}$/2.63\,mm, rotation no better than the prior. The trust
region is therefore load-bearing for our method's headline mean (castle's
raw twisted solution is 4.905$^{\circ}$/20.69\,mm), and no baseline in
Table~\ref{tab:mobilebrick} uses such a fallback. The classical control
needs none at this prior quality; Section~\ref{sec:exp-basin} shows the
ranking inverts as soon as the prior degrades.

A prior-free anchor completes the picture (lower block of
Table~\ref{tab:mobilebrick}, timings in Table~\ref{tab:runtime}).
GLOMAP~\citep{pan2024glomap}, run on the same feature databases without
the priors, registers every frame on all 18 scenes; on castle it
converges to the same twist as our solver
(Section~\ref{sec:exp-failure}). Prior-free global SfM is thus a strong
fallback when no prior exists, but on this benchmark it does not reach
the prior-seeded band, and its reconstruction lives in an arbitrary
similarity frame that must be aligned to be metric at all. The
feed-forward end of the prior-free spectrum, VGGT~\citep{wang2025vggt},
predicts all cameras of a sequence in one forward pass with every frame
registered, and lands well outside the prior-seeded band, behind the
ARKit prior itself; its currency here is speed and prior-freeness, not
accuracy. That currency composes with ours. Feeding VGGT's predicted
poses to our solver as the prior, under the same protocol as every
other row, improves all 18 scenes to 0.50$^{\circ}$/2.81\,mm,
castle-excluded 0.245$^{\circ}$/1.75\,mm, within ground-truth noise of
the ARKit-primed result; the plain solve suffices, since a
1.5$^{\circ}$ prior sits well inside its basin. A feed-forward front
end is then simply another coarse-prior source for classical
refinement, at CPU cost on top of the forward pass, consistent with the
match-graph bootstrap of Section~\ref{sec:exp-basin}. The composition
is not specific to one front-end family; MASt3R-SfM's output refines
identically on its subset, 1.26$^{\circ}$/5.84\,mm to
0.221$^{\circ}$/1.24\,mm, all six scenes improved. MASt3R-SfM~\citep{duisterhof2025mast3rsfm}, learned matching
with a sparse global alignment, poses every frame of a six-scene subset
and lands with VGGT rather than with GLOMAP. (This is a system-level
comparison; it changes matcher, initialization and optimizer at once,
so it does not isolate the objective difference discussed in
Section~\ref{sec:related}.) The pattern is consistent.
Prior-free systems, classical, learned or feed-forward, trade away the
metric anchor that the prior supplies for free, and none approaches the
prior-seeded band by itself; composed with our refiner as the prior
source, the feed-forward output does.

One hedge applies to all rankings here. MobileBrick's ground-truth poses
are themselves refined estimates with roughly $0.5^{\circ}$/2\,mm residual
accuracy~\citep{li2023mobilebrick}. Differences at or below that scale,
e.g.\ 0.218$^{\circ}$ vs.\ 0.22$^{\circ}$ or 1.50\,mm vs.\ 1.23\,mm, are
within ground-truth noise and should not be ranked. The defensible reading
of Table~\ref{tab:mobilebrick} is that prior-seeded refiners, classical
and learned, land in the same accuracy band, roughly $2\times$ better than
the prior in rotation.

\begin{table}[t]
  \centering
  \caption{Pose accuracy on the MobileBrick evaluation set, mean over
    scenes, lower is better. Baseline numbers from
    PoRF~\citep{bian2024porf}, Table~3 (the BARF, L2G-NeRF, and SPARF rows
    are PoRF's re-runs under its protocol). The right-hand block drops
    castle, the one scene whose matches prefer a wrong optimum for every
    filter-free solver tested (Section~\ref{sec:exp-failure}), and it is where
    the trust region stops mattering. Differences at or below the
    ${\sim}0.5^{\circ}$/2\,mm ground-truth accuracy are not rankable.
    Rotation in degrees, translation in millimetres.}
  \label{tab:mobilebrick}
  \small
  \setlength{\tabcolsep}{3pt}
  \begin{tabular}{lcccc}
    \toprule
    & \multicolumn{2}{c}{All 18 scenes} & \multicolumn{2}{c}{Excl.\ castle} \\
    \cmidrule(lr){2-3}\cmidrule(lr){4-5}
    Method & Rot. & Tr. & Rot. & Tr. \\
    \midrule
    ARKit prior~\citep{apple_arkit} & 0.46 & 1.90 & --- & --- \\
    BARF~\citep{lin2021barf}        & 0.57 & 2.50 & --- & --- \\
    L2G-NeRF~\citep{chen2023l2g}    & 0.58 & 2.84 & --- & --- \\
    SPARF~\citep{truong2023sparf}   & 0.61 & 3.39 & --- & --- \\
    PoRF~\citep{bian2024porf}       & 0.22 & 1.23 & --- & --- \\
    \midrule
    COLMAP BA from prior~\citep{schoenberger2016colmap} & 0.218 & 1.50 & 0.191 & 1.38 \\
    GLOMAP, prior-free~\citep{pan2024glomap}            & 0.61  & 2.96 & 0.36  & 1.91 \\
    VGGT, feed-forward~\citep{wang2025vggt}             & 1.51  & 6.97 & 1.29  & 6.17 \\
    VGGT prior + ours (plain)                           & 0.50  & 2.81 & 0.245 & 1.75 \\
    MASt3R-SfM~\citep{duisterhof2025mast3rsfm}$^{\ddagger}$ & --- & --- & 1.26 & 5.84 \\
    MASt3R prior + ours (plain)$^{\ddagger}$            & --- & --- & 0.221 & 1.24 \\
    \midrule
    Ours (full)$^{\dagger}$                & 0.246 & 1.61 & 0.228 & 1.57 \\
    Ours (full, no trust region)           & 0.488 & 2.63 & 0.228 & 1.57 \\
    \bottomrule
    \multicolumn{5}{@{}p{0.92\linewidth}@{}}{\footnotesize
      $^{\dagger}$18-scene mean with the prior trust region active; on one
      scene (castle) the solve falls back to the ARKit prior, which is
      scored as our result there (Section~\ref{sec:exp-failure}). The
      no-trust-region row scores castle's raw solution instead
      (4.905$^{\circ}$/20.69\,mm), which is the whole difference between
      the two rows. The COLMAP control uses no fallback.
      $^{\ddagger}$Six evaluation scenes only (aston, bridge, cabin,
      camera, jeep, satellite), none of them castle, so it is listed in
      the castle-free block; the subset differs from the other rows there.}
  \end{tabular}
\end{table}

\subsection{Room-scale refinement on ScanNet++}
\label{sec:exp-scannetpp}

MobileBrick's scenes are object orbits, so we replicate the campaign at
room scale on 15 iPhone-video scenes of the ScanNet++
\texttt{nvs\_sem\_val} split~\citep{yeshwanth2023scannetpp}. That split
holds 50 scenes spanning 13 scene types, and we took one scene of each
type, two of the two commonest (office and apartment), which fixes the
15 as a type-stratified cover of the pool rather than a sample of it.
The selection was made and frozen before any solver was run on this
benchmark, so no result could steer it. ScanNet++ registers a subsampled
set of video frames against a laser scan with COLMAP; we keep exactly
those frames (284--1071 per scene, 10{,}872 in total) and take the
shipped scan-aligned per-frame ARKit poses as the prior, which sits at
0.55$^{\circ}$/19\,mm scene-mean against the laser-registered ground
truth, ARKit-grade. That ground truth carries no published accuracy of
its own. The release renders pseudo-images from the laser scan into a
COLMAP reconstruction of the video frames, refines the resulting poses
photometrically against the scan geometry, and discards frames whose
iPhone depth disagrees with the rendered scan depth by more than
0.3\,m on average, but it reports no pose residual against the scan, so
the benchmark's own floor is unquantified rather than
small~\citep{yeshwanth2023scannetpp}. Two consequences follow and we
state both. Pose differences below that unknown floor are not
measurable, which is how the plain solve's 0.57$^{\circ}$ against the
prior's 0.55$^{\circ}$ should be read, and the ground truth is itself
COLMAP output, so evaluating COLMAP-style refinement against it carries
a circularity risk we cannot bound from the published information. What
we claim from the nominal room-scale result is the direction of the
classical arm's move, consistent in all 15 scenes, not the size of the
0.19$^{\circ}$ gap it opens. Each scene gets one shared SIFT database,
so all solvers see identical matches. One protocol change is forced by the
data. The video frames carry real lens distortion that the pinhole ARKit
intrinsics do not model, so our arms also self-calibrate (focal length,
principal point and distortion alongside poses), and the classical
control's bundle adjustment runs the OPENCV camera model with focal,
principal-point and distortion refinement enabled for the same freedom.
How that freedom is scheduled matters at room scale. The plain solve
couples self-calibration with the pose solve under an orientation prior
of $\lambda_R = 10^{3}$, fine at zero perturbation. The GNC arm stages
the two, a poses-only annealed pass from the perturbed prior under a
light orientation anchor ($\lambda_R = 10^{1}$, matches capped at 100
per pair), then one warm self-calibration pass re-anchored at the
pass-one poses; annealing the coupled solve instead lets the freed
intrinsics absorb exactly the pose error the annealing is trying to
remove, and we quantify the difference below. The freed intrinsics track
physical calibration rather than absorbing pose error; on the office
development scene the self-calibrated solve independently recovers the
GT calibration, focal (1436.8, 1440.7) against COLMAP's (1435.3,
1438.4) and distortion $k_1$/$k_2$ 0.074/$-0.090$ against
0.069/$-0.084$. The perturbation protocol mirrors
Section~\ref{sec:exp-basin}, i.i.d.\ $1^{\circ}$--$32^{\circ}$ doubling
with translation at 5\,mm per degree, 735 runs in total. The classical
control and the coupled GNC variant carry two seeds, 195 runs each
(six perturbation levels at 30 runs plus the 15 zero-perturbation
solves); the staged GNC variant carries five seeds from $8^{\circ}$
upward, 330 runs; the plain solve runs at zero perturbation only, 15.
None of the 735 crashed or diverged to a
solver error; ``failure'' in the rest of this section always means a
completed solve that fails to recover the prior, never a broken one.

At zero perturbation the classical control degrades the prior in 15 of
15 scenes (scene-mean rotation 0.55$^{\circ} \rightarrow 0.74^{\circ}$,
translation 19 $\rightarrow$ 21\,mm). Tuning does not rescue it; the
deep loose-gate variant of the object-scale envelope still degrades 12
of 15 scenes (0.55$^{\circ} \rightarrow 0.70^{\circ}$ mean) at ten
times the cost. This is the object-scale
mechanism operating on a good prior; triangulating from ARKit-grade
poses at room distances commits enough wrong structure that the polish
moves every scene the wrong way. The plain solve holds the prior
instead (0.57$^{\circ}$/18.5\,mm scene-mean against the prior's
0.55$^{\circ}$/19\,mm), consistent with priors already
near the benchmark's own registration floor, and the staged GNC recipe
pays a measurable nominal price for its basin
(0.69$^{\circ}$/19.6\,mm scene-mean at a 0.52$^{\circ}$ median); the two scenes whose
priors exceed $1^{\circ}$ (a kitchen and a lab) stay above $1^{\circ}$
under every solver, and the staged recipe adds one more, an apartment
whose solve settles at a ${\sim}1.1^{\circ}$ floor at every
perturbation level, the nominal price of the light pass-one anchor.
Cost scales with the problem size, a median of 14\,CPU-minutes per
scene for the staged GNC solve against 1.7 minutes for the classical
control.

Table~\ref{tab:snpp} sweeps the perturbation ladder. The classical
control recovers nothing at any level; its median tracks the
perturbation and it halves the prior's rotation error in 2 of 180 runs
across the whole sweep. The staged GNC arm is flat. Its median is
0.52$^{\circ}$--0.54$^{\circ}$ at every level through
$16^{\circ}$/80\,mm and
0.59$^{\circ}$ at $32^{\circ}$/160\,mm, with recovered translation
(16\,mm median at every level) better than the unperturbed prior's own
19\,mm, no
solver failure in 330 runs, and the prior's rotation error at least
halved in every run from $4^{\circ}$ upward. Almost all runs above
$1^{\circ}$ are the same kitchen, lab and apartment whose nominal
solves sit at or above that mark, at every one of the five seeds, with
a copy room joining them marginally from $8^{\circ}$; the perturbation
barely adds to those, the lab's worst reaching 2.4$^{\circ}$ at
$32^{\circ}$. The tail holds one genuine excursion. A second apartment,
clean at every level through $16^{\circ}$ and at four of five seeds at
$32^{\circ}$, lands at 6.5$^{\circ}$ on the fifth, the one
seed-dependent room-scale failure the five-seed sweep exposes and the
two-seed sweep of an object-scale-sized budget would have missed.
Staging is what
buys this width. The coupled single-stage variant recovers only
through $4^{\circ}$/20\,mm (median 0.47$^{\circ}$--0.73$^{\circ}$,
20--29 of 30 runs halving the prior), is partial at $8^{\circ}$
(median 2.5$^{\circ}$) and lost from $16^{\circ}$ (median
14$^{\circ}$), its freed intrinsics absorbing the very pose error the
annealing removes; where it fails it wanders in translation rather
than returning near the prior, and a factor-two test on Sim(3)-aligned
median per-frame displacement against the prior, the room-scale
analogue of the trust region of Section~\ref{sec:variants}, flags 28
of its 37 ladder failures with 5 false alarms in 104 healthy solves
(falling back to the prior on a flag cuts its worst $16^{\circ}$
median from 657 to 82\,mm). The staged recipe leaves nothing for that
guardrail to catch through $32^{\circ}$.

The two campaigns together sharpen the paper's claim. The room-scale
basin, measured on the staged recipe, extends through
$32^{\circ}$/160\,mm at a median indistinguishable from nominal,
matching the object-scale grid of Section~\ref{sec:exp-basin}; the $8^{\circ}$--$16^{\circ}$ edge of
the coupled solve was a property of annealing through free intrinsics,
not of the scene scale. The basin's ratio to the classical basin
exceeds an order of magnitude at both scales, and the mechanism that
widens it is the same one that makes refinement safe to run when the
prior is already good. Whether to run the annealed solve therefore
does not depend on prior quality; it holds ARKit-grade priors that
classical refinement reliably damages at room scale, and repairs
perturbed priors that classical refinement cannot. What does depend on
prior quality is the guardrail configuration, and the conclusion
assembles the pre-solve and posterior checks measured across the
campaigns into an explicit recipe for it.

\begin{table}[t]
  \centering
  \caption{Room-scale basin on 15 ScanNet++ scenes, two seeds per level
    (30 runs per level and solver), i.i.d.\ perturbations, protocol of
    Section~\ref{sec:exp-basin} with self-calibration for our arm
    (staged after a poses-only annealed pass) and the OPENCV model for
    the classical control. ``med.'' columns are
    the median final rotation (deg) and translation (mm),
    ``${<}1^{\circ}$'' counts runs below $1^{\circ}$ absolute error
    (the misses are the kitchen, lab and apartment whose
    nominal solves already sit at or above $1^{\circ}$ at every seed,
    joined marginally by a copy room from $8^{\circ}$),
    ``$\times\tfrac12$'' counts runs that at least halve the prior's
    rotation error. Our arm carries two seeds below $8^{\circ}$ and five
    at $8^{\circ}$ and above, so its counts are out of 30 and out of 75
    respectively; the classical control carries two seeds throughout, 30
    runs per level. Each perturbation level pairs the stated rotation
    with 5\,mm per degree.}
  \label{tab:snpp}
  \footnotesize
  \setlength{\tabcolsep}{3pt}
  \begin{tabular}{l rrrr rrrr}
    \toprule
    & \multicolumn{4}{c}{Classical tri.+BA}
      & \multicolumn{4}{c}{Ours, staged GNC} \\
    \cmidrule(lr){2-5} \cmidrule(lr){6-9}
    Pert. & med.$^{\circ}$ & mm & ${<}1^{\circ}$ &
    $\times\tfrac12$ & med.$^{\circ}$ & mm & ${<}1^{\circ}$ &
    $\times\tfrac12$ \\
    \midrule
    $1^{\circ}$    & 1.96 & 56 & 6 & 2 & \textbf{0.52} &
    \textbf{16} & \textbf{24} & \textbf{18} \\
    $2^{\circ}$   & 2.47 & 66 & 0 & 0 & \textbf{0.52} &
    \textbf{16} & \textbf{24} & \textbf{25} \\
    $4^{\circ}$   & 3.87 & 70 & 0 & 0 & \textbf{0.52} &
    \textbf{16} & \textbf{24} & \textbf{30} \\
    $8^{\circ}$   & 7.88 & 93 & 0 & 0 & \textbf{0.52} &
    \textbf{16} & \textbf{59} & \textbf{75} \\
    $16^{\circ}$  & 15.86 & 142 & 0 & 0 & \textbf{0.54} &
    \textbf{16} & \textbf{57} & \textbf{75} \\
    $32^{\circ}$ & 31.59 & 269 & 0 & 0 & \textbf{0.59} &
    \textbf{16} & \textbf{54} & \textbf{75} \\
    \bottomrule
  \end{tabular}
\end{table}

\subsection{Runtime}
\label{sec:exp-runtime}

Table~\ref{tab:runtime} compares wall-clock cost per scene. The
learning-based refiners train a per-scene network on GPU; the classical
control and our method are sparse solves on a single CPU (the Ryzen~9
9950X of Section~\ref{sec:exp-setup}, 32 threads). All CPU figures cover the solve only. Feature
extraction and matching are excluded, as all listed methods presuppose
correspondences or an equivalent front-end. We report the full spread
rather than a median alone, since scene size drives an order of
magnitude of variation within every CPU method (big\_ben is the worst
case throughout). The GNC schedule of Section~\ref{sec:gnc} multiplies
solve cost by about $2\times$ (median paired ratio 1.88 over the matched
basin runs; warm starts make the three stages much cheaper than three
plain solves). Extraction and matching are excluded symmetrically
rather than by convention alone. Every arm in every campaign reads the
same per-scene database, built once by exhaustive SIFT matching, and
every clock starts after that database is in place, so no arm hides a
front-end cost in another arm's total.

Table~\ref{tab:runtime} reports nominal configurations, which is the
wrong comparison for a coarse prior, since the arms that survive one
are not the arms that are cheap at nominal. Measured on the same
machine under the same load over the six control scenes, the ordering
at coarse priors, all medians over those same six scenes, is ours at
35\,s, prior-position mapping at 45\,s and annealed re-triangulation at
87\,s, with modern default
bundle adjustment far cheaper at 13\,s but dead past
$1$--$2^{\circ}$, and PoRF at 1.8--4.6 GPU-hours per run without
recovering any of them (Section~\ref{sec:exp-basin}). Among methods
that reach a coarse prior at all, the structureless solve is the
cheapest at object scale. At room scale the ordering changes and we
report the loss plainly. Our staged two-pass recipe runs at 780--920\,s
per scene against 1031\,s for the tuned deep loose-gate classical
control, which runs at 1.7 minutes at its defaults
(Section~\ref{sec:exp-scannetpp}), and 360\,s for
prior-position mapping, so re-mapping is roughly twice as fast as
refining a room, and it is the only arm anywhere in this evaluation
that is both faster than ours and able to handle a coarse prior. It
achieves that by discarding the prior, at the cost quantified in
Section~\ref{sec:exp-frame}.

The comparison to the learned refiners needs care, and we avoid a single
``$X\times$ less compute'' figure. PoRF reports 2.5\,h per scene for its
full 50k-iteration training on an NVIDIA A40~\citep{bian2024porf}. That
budget also produces a surface reconstruction (a NeuS field) in addition
to poses, and PoRF notes that pose accuracy typically converges within the
first 5{,}000 iterations, so its effective pose-refinement cost is well
below 2.5\,h. What survives these caveats is that on this benchmark,
prior-seeded sparse bundle adjustment refines poses in CPU-seconds to
CPU-minutes of wall-clock, versus GPU-minutes to GPU-hours for the learned
refiners, on different hardware and with different by-products.

\begin{table}[t]
  \centering
  \caption{Per-scene runtime on the MobileBrick evaluation scenes, in
    seconds unless marked. Baseline times are the published
    figures~\citep{bian2024porf,truong2023sparf}; CPU times cover the
    solve only (matching excluded) on a single CPU (32 threads), GPU
    times the full inference or alignment.}
  \label{tab:runtime}
  \small
  \setlength{\tabcolsep}{1.5pt}
  \begin{tabular}{llcccc}
    \toprule
    Method & Hardware & Min & Med. & Mean & Max \\
    \midrule
    PoRF~\citep{bian2024porf}     & A40 GPU  & \multicolumn{4}{c}{2.5\,h$^{\S}$} \\
    SPARF~\citep{truong2023sparf}$^{\ddagger}$ & A100 GPU & \multicolumn{4}{c}{${\sim}$10\,h} \\
    \midrule
    COLMAP BA from prior          & 1 CPU & 9 & 53  & 55  & 139 \\
    Ours (full)                   & 1 CPU & 9 & 101 & 141 & 580 \\
    Ours (cap 100)                & 1 CPU & 1 & 10 & 11 & 36 \\
    \midrule
    GLOMAP, prior-free            & 1 CPU & 19 & 90 & 94 & 275 \\
    MASt3R-SfM$^{*}$              & 5090 GPU & 80 & 90 & 105 & 152 \\
    VGGT, feed-forward            & 5090 GPU & 3  & 8  & 10 & 64 \\
    \bottomrule
    \multicolumn{6}{@{}p{0.92\linewidth}@{}}{\footnotesize
      $^{\S}$Full 50k-iteration training, which also produces a surface
      reconstruction; PoRF reports pose convergence typically within the
      first 5{,}000 iterations.
      $^{\ddagger}$SPARF's published timing is for its own sparse 3-view
      setting; no full-sequence timing is reported.
      $^{*}$Six-scene subset, as in Table~\ref{tab:mobilebrick}.}
  \end{tabular}
\end{table}

\subsection{Downstream novel-view synthesis}
\label{sec:exp-gsplat}

Pose error matters only through its downstream effect, so we close the
loop with 3D Gaussian splatting~\citep{kerbl20233dgs}. On all 18
evaluation scenes we train a
splatting model per pose source with every 8th frame held out, 7{,}000
steps at half resolution, poses held fixed, and initialization from a
random point sphere rather than an SfM point cloud so that no pose
source is favored; the Gaussian budget and all other hyperparameters
are identical across sources. Table~\ref{tab:gsplat} scores held-out
views for four pose sources, the raw ARKit prior, our refined poses
(full), the COLMAP-BA control and the dataset ground truth.

Refinement recovers one half (ours) to two thirds (the control) of the
ARKit-to-ground-truth PSNR gap. The 0.23\,dB mean gap between our poses
and the control is castle in disguise; castle is scored as the returned
prior by design (Section~\ref{sec:exp-failure}), and excluding it the
gap is 0.10\,dB, consistent with the not-rankable band of
Table~\ref{tab:mobilebrick}. A training-noise floor calibrates these
numbers. On castle our scored poses and the prior are the same pose
set, and their independently trained models still differ by 0.39\,dB,
so per-scene differences below ${\sim}0.4$\,dB carry no signal.
Per-scene results are in the appendix
(Table~\ref{tab:gsplat-perscene}). On five scenes for ours and six for
the control the refined poses render
better than the ground-truth poses do, consistent with pose
differences at this scale sitting at the benchmark's ground-truth noise
floor. Rendering alone cannot certify that reading, since castle's
twisted solve also out-renders the ground truth
(Section~\ref{sec:exp-failure}); the sub-noise-floor pose gaps of
Table~\ref{tab:mobilebrick} are what carry it.

\begin{table}[t]
  \centering
  \caption{Downstream 3D Gaussian splatting on held-out views, mean over
    the 18 evaluation scenes. Identical training configuration per pose
    source; poses held fixed.}
  \label{tab:gsplat}
  \small
  \setlength{\tabcolsep}{4pt}
  \begin{tabular}{lccc}
    \toprule
    Pose source & PSNR $\uparrow$ & SSIM $\uparrow$ & LPIPS $\downarrow$ \\
    \midrule
    ARKit prior          & 16.55 & 0.554 & 0.389 \\
    Ours (full)          & 17.48 & 0.659 & 0.337 \\
    COLMAP BA from prior & 17.71 & 0.667 & 0.331 \\
    Ground truth         & 18.29 & 0.702 & 0.314 \\
    \bottomrule
  \end{tabular}
\end{table}

\subsection{Ablation of the parameterization and match capping}
\label{sec:exp-ablation}

To isolate the effect of the parameterization, we compare the full
per-observation formulation against the anchored-track variant of
Section~\ref{sec:variants} (same matches, same priors, same robust loss,
differing only in the structure variables) and against the per-observation
formulation with matches capped at 100 per image pair
(Table~\ref{tab:ablation}). The mean differences here sit below the
ground-truth noise floor of Section~\ref{sec:exp-mobilebrick} and are
rankable only in the paired per-scene sense, where the shared
ground-truth error partially cancels; we therefore report per-scene win
counts alongside the means. We also ablate the orientation-prior default;
with $\lambda_R = 10^3$ vs.\ $0$, translation improves and rotation is
flat (better on 13 and 12 of 18 scenes respectively), a small but
consistent win. The trust region falls back
on castle in every mode. Anchored tracks are slightly worse though
faster (the full formulation is better on 14 of 18 scenes in
translation, 11 in rotation); under GNC they share the wide basin
(Section~\ref{sec:exp-basin}).
A plausible explanation, consistent with MobileBrick's measured per-frame
focal drift (0.9--2.9\% per scene) though we have not isolated the
mechanism experimentally, is that the per-observation slack absorbs
unmodeled per-frame intrinsics variation that rigid tracks force into the
poses. Capping the matches is the best speed/accuracy trade and makes our
solver the fastest variant evaluated in this paper. The cap draws its
budget uniformly at random per pair, seeded by the pair's image ids so
the draw is reproducible; selecting the same budget for coverage or
pose information instead, in the spirit of view-graph
selection~\citep{sweeney2015viewgraph}, is an untested refinement.

\begin{table}[t]
  \centering
  \caption{Ablation on the MobileBrick evaluation set (18 scenes), mean
    pose error and median solve time.}
  \label{tab:ablation}
  \small
  \setlength{\tabcolsep}{2.5pt}
  \begin{tabular}{lccc}
    \toprule
    Variant & Rot.\ ($^{\circ}$) $\downarrow$ & Trans.\ (mm) $\downarrow$
      & Time \\
    \midrule
    Per-obs.\ (full, $\lambda_R{=}10^3$) & 0.246 & 1.61 & 101\,s \\
    Per-obs.\ (full, $\lambda_R{=}0$)    & 0.245 & 1.66 & 98\,s \\
    Anchored tracks ($\lambda_R{=}0$)    & 0.277 & 1.82 & 25\,s \\
    Per-obs., cap 100 ($\lambda_R{=}0$)  & 0.265 & 1.80 & 10\,s \\
    \bottomrule
  \end{tabular}
\end{table}

\subsection{Why triangulation narrows the basin}
\label{sec:exp-mechanism}

\begin{figure*}[t]
  \centering
  \includegraphics[width=0.85\textwidth]{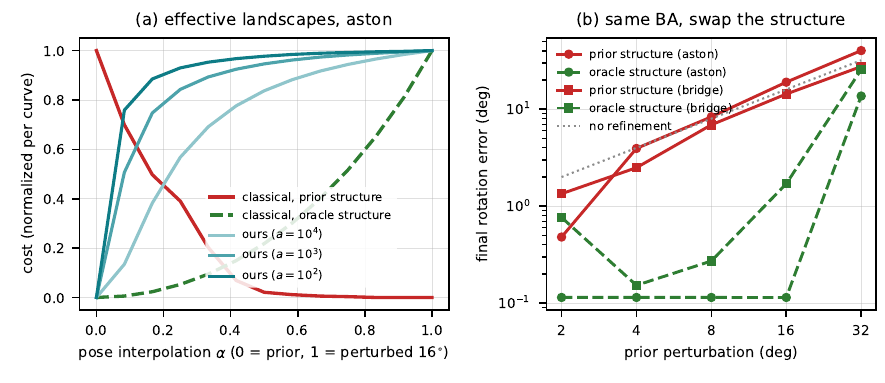}
  \caption{Why triangulation narrows the basin. \textbf{(a)}
    Measured cost along the pose interpolation from the accurate prior
    ($\alpha{=}0$) to a $16^{\circ}$/80\,mm perturbation ($\alpha{=}1$) on
    aston, each curve min--max normalized (only shape and argmin are
    comparable). Classical BA with structure triangulated at the perturbed
    prior and held fixed has its minimum at the perturbation. The same
    objective with oracle structure, and our structure-marginalized
    objective at every arctan scale, have theirs at the accurate prior,
    with larger scales retaining gradient along the whole path (the
    annealed landscapes GNC descends). \textbf{(b)} The causal control.
    Identical COLMAP BA from identical perturbed poses, swapping only the
    triangulation input (seed 0). Oracle structure recovers through
    $16^{\circ}$ where prior structure collapses from $2^{\circ}$; both
    fail at $32^{\circ}$, where GNC still recovers
    (Figure~\ref{fig:basin}).}
  \label{fig:concept}
\end{figure*}

Before measuring basins, we isolate the mechanism sketched in
Figure~\ref{fig:teaser}a. Triangulating from the prior is a commitment.
\texttt{point\_triangulator} retains only observations consistent with the
given poses (on aston at a $16^{\circ}$ perturbation, 52{,}814
observations survive, against 580{,}183 at the accurate prior, i.e.\ 9\%),
and the retained structure then defines which poses look good.
Figure~\ref{fig:concept}a measures this directly as one-dimensional cost
landscapes along the geodesic interpolation from the accurate prior
($\alpha{=}0$) to its perturbed copy ($\alpha{=}1$). With structure
triangulated at the perturbed prior and held fixed, the reprojection cost
decreases monotonically toward the perturbation; the wrong initialization
is the objective's minimum, drawn. Swapping in oracle structure
(triangulated at the accurate prior) flips the same objective's minimum to
$\alpha{=}0$, so the poison is the triangulation input rather than the
objective form. Our marginalized objective, with poses frozen and ray
times solved out, which is exact since the depths are mutually independent
given poses, has its minimum at $\alpha{=}0$ at every loss scale. The
annealed scales show the GNC effect on the actual landscape. At
$a{=}10^2$ the cost saturates within a few percent of the interpolation
(sharp minimum, flat far field), while $a{=}10^4$ retains gradient along
the entire path.

Figure~\ref{fig:concept}b makes the claim causal. The same COLMAP bundle
adjustment runs from the same perturbed poses with only the structure
input swapped. With oracle structure it recovers through $16^{\circ}$
(aston, $0.11^{\circ}$ at every level) where the prior-structure pipeline
collapses from $2^{\circ}$. Two readings follow. First, classical BA's
basin collapse is entirely a property of triangulation from the prior.
Second, even oracle structure fails at $32^{\circ}$, where the annealed
structureless objective still recovers (Section~\ref{sec:exp-basin});
in this regime the annealed re-expressed structure is stronger than
committing even perfect structure once (the oracle arm runs the nominal
loss, so the comparison bundles re-expression with annealing).

\subsection{Basin of convergence}
\label{sec:exp-basin}

\begin{figure}[t]
  \centering
  \includegraphics[width=\linewidth]{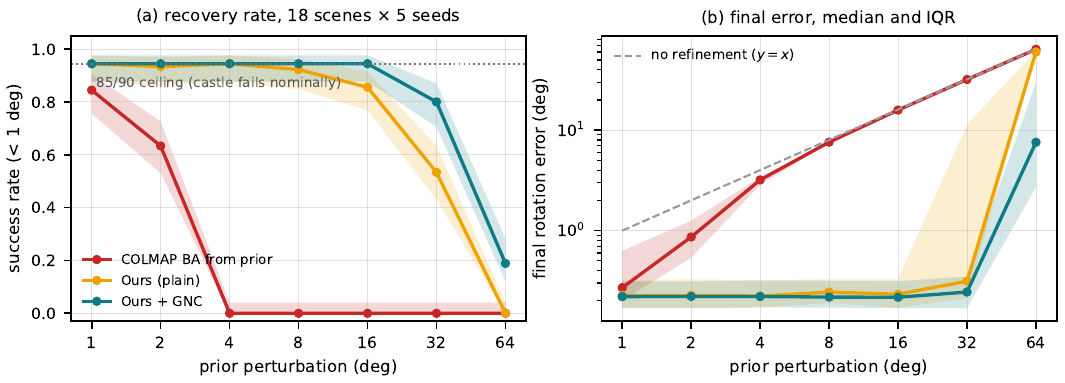}
  \caption{Basin of convergence for three solvers started from identical
    perturbed priors, 18 scenes with 5 noise seeds per level (630 runs
    per solver; trust region disabled). \textbf{(a)} Fraction of runs
    recovering below $1^{\circ}$ final rotation error, with Wilson 95\%
    bands. The dotted ceiling (85/90) is castle, whose repeated-texture
    failure is nominal, not basin-induced
    (Section~\ref{sec:exp-failure}). Classical COLMAP BA (red) collapses
    to zero from $4^{\circ}$/20\,mm, since triangulation from a coarse
    prior leaves nothing to optimize; the triangulation-free solve
    (orange) degrades stochastically from $2^{\circ}$ and materially
    from $16^{\circ}$; GNC (teal)
    recovers every non-castle run through $16^{\circ}$/80\,mm and 72/90
    at $32^{\circ}$/160\,mm (85\% of the non-castle runs). \textbf{(b)} Median final rotation error
    with interquartile band (log--log); the dashed line means returning
    the perturbed prior unchanged. Where recovery happens, it is to
    nominal accuracy (${\sim}0.22^{\circ}$).}
  \label{fig:basin}
\end{figure}

This experiment measures, at its sharpest, the defect diagnosed in Section~\ref{sec:exp-scannetpp}. We perturb the pose
priors with paired i.i.d.\ per-frame rotation/translation noise of
increasing magnitude, $1^{\circ}$/5\,mm doubling up to
$64^{\circ}$/320\,mm with five noise seeds per level, on all 18
evaluation scenes (cap 100), and run all three solvers from the same
perturbed priors, 630 runs per solver. These are the classical COLMAP
control, our plain solve, and our solve with the GNC schedule of
Section~\ref{sec:gnc} (Figure~\ref{fig:basin}). All three arms read one
shared per-scene SIFT database, and the cap is ours alone. Our solver
subsamples each image pair to 100 matches deterministically as it loads
the database into the problem, while the classical control is handed the
database file itself and its triangulator consumes every stored two-view
correspondence. The asymmetry therefore runs against us; the control
sees strictly more evidence per pair than our capped arms do, so the
basin gap we report is not bought by starving it. Uncapping our side
moves mean rotation error from 0.265$^{\circ}$ to 0.245$^{\circ}$ at ten
times the solve time (Section~\ref{sec:exp-ablation}), which is why the
ladder runs capped.
We call a run a success
when its final mean rotation error is below $1^{\circ}$, four times the
nominal accuracy and no larger than the smallest perturbation level;
intervals are Wilson 95\% scores. The rotation-only form costs nothing;
every run that either arm of our solver counts as a
success also lands below 8\,mm mean aligned translation error, with a
single exception among the ten seeds at $32^{\circ}$ where a
colosseum solve reaches 0.92$^{\circ}$ but 17.9\,mm, so a joint
criterion would change exactly one count in 1345. The prior trust region is disabled throughout,
for a structural reason. The synthetic perturbations exceed its envelope
by construction, so the guardrail would reject exactly the corrections the
experiment measures; the experiment probes each objective's basin, not the
shipped configuration. (Running the shipped configuration anyway, with
full matches, orientation prior and the trust region at
$3.5^{\circ}$/22\,mm, from $4^{\circ}$ and $8^{\circ}$ perturbed priors
trips the guardrail in 24 of 24 runs and returns the perturbed prior,
as designed; a correct recovery must move the poses further than the
envelope allows. When the prior is suspected coarse the trust region
must be widened or disabled.)

The two configurations differ in more than the guardrail, so we
measured what the shipped one costs on the ladder itself. Running the
headline configuration, full matches and the orientation prior at
$10^{3}$, over the six control scenes at 2 to $32^{\circ}$ with three
seeds, and gating each solve afterwards with an envelope widened by
the declared perturbation level, it recovers 18 of 18 runs at
$2^{\circ}$ and $4^{\circ}$, 15 of 18 at $8^{\circ}$, and none at
$16^{\circ}$ or $32^{\circ}$ (medians 4.9$^{\circ}$ and
30.4$^{\circ}$), where the basin configuration recovers 18 of 18 at
every one of those levels. Widening the envelope is therefore only
the outer half of the cost. The rest is the shipped weights
themselves; anchoring rotations to a prior that is wrong by
$16^{\circ}$ holds the solve near that prior, which is the correct
behavior for the regime those weights were chosen for and the wrong
one here. The basin we report is a property of the objective under a
coarse-prior configuration, not of a single fixed deployment, and the
recipe of Section~\ref{sec:conclusion} is what selects between them.

Table~\ref{tab:basin} reports the counts and Figure~\ref{fig:basin}
plots them with Wilson bands. Classical BA reproduces its nominal
per-scene accuracy at zero perturbation (0.118--0.150$^{\circ}$ on the
three basin development scenes, aston, bridge and camera) and then
collapses almost immediately. From
$4^{\circ}$/20\,mm onward it succeeds in no run at
any level (upper 95\% bound 4\%), returning the perturbed prior
essentially unchanged, which is what its median error tracking the
perturbation level means. The mechanism is visible in the solver itself.
\texttt{point\_triangulator} retains only matches consistent with the
given poses, so a coarse prior yields a sparse, wrong point cloud
(bridge's solve time drops from ${\sim}8$\,s to ${\sim}4$\,s because
almost nothing survives triangulation), and bundle adjustment then
polishes that wrong structure. ARKit's real error (${\sim}0.46^{\circ}$)
sits just inside this narrow basin, which is exactly why the classical
control excels in Table~\ref{tab:mobilebrick}.

The collapse threshold is a property of the configuration, so we also
report the control at its best settings rather than only its defaults.
The control follows COLMAP's documented
reconstruct-from-known-poses recipe~\citep{schoenberger2016colmap},
and its knob search ran on the same six control scenes as our schedule
development, deepening the triangulate-adjust alternation from two to
eight rounds and loosening the triangulation gates (create and
continue angle errors to $8^{\circ}$, merge, complete and filter
reprojection errors to 16\,px). The tuned control is genuinely
stronger. Re-triangulation after each adjustment re-admits matches the
previous gate culled, and the deep loose variant recovers 17 of 18
runs at $4^{\circ}$ and at $8^{\circ}$/40\,mm (medians
$0.17^{\circ}$--$0.26^{\circ}$), a regime the default is dead in, at
five times the cost (median 102\,s against 20\,s). The envelope still
ends well inside ours. At $16^{\circ}$/80\,mm the best classical
configuration recovers 1 of 18 runs (3 of 18 under kidnap error) and
none at $32^{\circ}$ in either mode, where GNC recovers every feasible
run at $16^{\circ}$ and 85\% at $32^{\circ}$ at a third of the tuned
control's cost.

Because the 3.7 build used throughout predates COLMAP's pose-prior
mapper and exposes no robust-loss option in its bundle adjuster, we
repeated the envelope on the current build (4.2.0.dev0, official
image). The version bump alone changes nothing; the modern defaults
reproduce the 3.7 collapse with identical sub-$1^{\circ}$ counts and
medians within a few percent at every level. The interesting arms are
the ones the CLI still does not expose. Cauchy-loss bundle adjustment
(scale 4\,px, via pycolmap) under the loose gates and eight rounds
recovers 15 of 18 runs at $16^{\circ}$ and 2 of 18 at $32^{\circ}$.
Porting our schedule wholesale, three stages of progressively
tightening triangulation gates and robust-loss scales with
warm-started poses and a fresh triangulation per stage, recovers 14 of
18 at $16^{\circ}$ (median $0.17^{\circ}$) and 7 of 18 at $32^{\circ}$
(median $3.7^{\circ}$), at a mean 100\,s per run; the residual misses
are near-$180^{\circ}$ mirror flips at the same seeds that fail in
the 3.7 campaign, and each arm ran exactly one configuration, so
these numbers are first-attempt, not tuned. Annealed re-expression is
therefore portable; re-triangulating per stage is a coarser, costlier
form of the mechanism our objective applies at every iterate, and it
carries classical refinement to $16^{\circ}$, no longer
deterministically. No classical refinement arm survives $32^{\circ}$.
Separately, the modern \texttt{pose\_prior\_mapper} solves every
level including $32^{\circ}$ (18 of 18, median $0.16^{\circ}$), and
it deserves a precise account because it is the strongest classical
baseline we found. It consumes position priors only and re-runs full
incremental SfM. We looked for the error structure that would break
it, expecting correlated relocalization error to poison a
position-prior term where independent jitter averages out, and found
none. Under kidnap perturbations it recovers 90 of 90 runs with every
frame registered, and its per-scene result varies by at most
0.002$^{\circ}$ across every level, seed and noise mode, matching its
i.i.d.\ values; a probe that scatters the priors by a metre while
declaring millimetre confidence returns the same poses again. The
prior changes its runtime and nothing else. Under Sim(3) scoring that
is the whole story, and the distinction is functional. Re-mapping does
not refine the given poses, it replaces them, so it cannot exploit a
prior that is already good, cannot be warm-started, and needs the full
image set and match graph on every run, whereas the whole premise here
is a prior worth keeping. Scored in the prior's own frame the two are
indistinguishable at every coarse level, so nothing is forfeited by
refining, and nothing is gained either
(Section~\ref{sec:exp-frame}). Reliability differs
too. On three ScanNet++ rooms it matches our staged arm's accuracy at
roughly half its cost on the two larger scenes, but on the third,
three of eight cells either crashed the mapper or returned a
globally bent model scored at 33.5$^{\circ}$ whose local pose
relations looked healthy, a failure with no cheap signal, where our
staged arm completed 330 of 330 room runs without one. The honest
summary is that gate-and-depth tuning buys classical refinement
roughly $8^{\circ}$, porting the annealing buys $16^{\circ}$, and
only the structureless solve refines through $32^{\circ}$; if
discarding the prior and re-mapping from scratch is admissible for
the application, prior-position mapping covers every level and should
be the baseline of choice, at accuracy indistinguishable from ours in
the prior's own frame.

These envelopes rest on unequal search budgets, which we tally for
symmetry. The classical side searched eight configurations, a
two-by-two grid of alternation depth and gate looseness on the 3.7
build plus one first-attempt configuration per modern arm with no
retune at any point; the room-scale control
(Section~\ref{sec:exp-scannetpp}) reran the object-scale default and
tuned settings unchanged. Our side searched more. The recorded
development loop spent fourteen configuration iterations on the six
development scenes, one knob change per iteration, covering three loss
scales, four orientation-prior weights and assorted staging,
initialization and convergence variants, and adopted two changes, the
orientation prior at $10^3$ and the GNC schedule itself, which was the
first schedule tried. Six alternative schedules run since, the
stage-count and start-scale variants of
Table~\ref{tab:basin-controls} and the higher or steeper starts probed
at the basin edge and at room scale, matched or lost to it, and the
room-scale staging was selected by a further six-arm grid on three
development rooms, with the wander threshold read off existing runs at
two candidate factors and no re-solving. The count is therefore
twenty-six solver configurations on our side against eight classical
ones, both searches run with sight of the same six control scenes, an
asymmetry of roughly three to one in our favor. What the search bought
differs more than its size. The classical envelope moves by a factor
of four in perturbation level between its default and tuned settings,
whereas ours is insensitive to everything we varied except a start
scale a decade too low; the width attaches to the formulation, not to
a tuned operating point.

The triangulation-free objective is wide but stochastic at the edge.
Castle's repeated-texture optimum defeats the solver already at zero
perturbation (Section~\ref{sec:exp-failure}), so 85 of 90 is the ceiling
for every arm of our solver here and the plain solve reaches it at
$1^{\circ}$, $4^{\circ}$ and nowhere else. Its shortfalls at
$2^{\circ}$, $8^{\circ}$ and $16^{\circ}$ are seed-dependent; at
$32^{\circ}$ they become scene-correlated.

GNC covers the wide basin without a failure through $16^{\circ}$/80\,mm,
about $35\times$ ARKit's real error. Every non-castle run succeeds at
every level up to $16^{\circ}$/80\,mm, 425 of 425 across 17 scenes and
five seeds (per-level Wilson interval $[0.96, 1.00]$ over runs; because
failures cluster by scene, the conservative scene-level reading is 17 of
17 scenes with Wilson interval $[0.82, 1.00]$), which eliminates
every seed-dependent failure of the plain solve, at unchanged nominal
accuracy and ${\sim}2\times$ solve cost (the paired ratio falls toward
$1\times$ at the extreme levels, where the plain solve's own median
cost doubles while the schedule's stays flat). At $32^{\circ}$/160\,mm,
about $70\times$ ARKit's real error, GNC recovers 85\% of feasible runs
(72/85, or 80\% of all 90) against
the plain solve's 56\%, the residual failures concentrating in four
scenes (beetles, space\_shuttle, colosseum, london\_bus). Because that
level carries the widest interval, we ran five further seeds there.
The rate is unchanged, 72 of 85 again on the new seeds and 144 of 170
over all ten (84.7\%, Wilson $[0.79, 0.89]$, narrowed from
$[0.76, 0.91]$), and the failures stay in the same scenes, twelve of
seventeen clean at every one of the ten seeds. At
$64^{\circ}$/320\,mm it still recovers runs where the plain solve
recovers none, but is no longer reliable. Probe runs between the grid
levels recover aston from $40^{\circ}$ and from $48^{\circ}$/240\,mm on
two seeds and porsche from $48^{\circ}$/240\,mm, so on scenes with
healthy match graphs the cliff sits between $48^{\circ}$ and
$64^{\circ}$. The four edge failures are diagnosable before solving
and repairable without a better prior. They are exactly the four
scenes whose match graphs carry the heaviest wrong-lobe population.
Ranking all 18 scenes by mean pairwise rotation-averaging residual, a
statistic computed from the feature database alone with no prior,
ground truth or solve entering it, the four occupy ranks 1 through 4
(10.8--23.0$^{\circ}$) strictly above every scene that recovers (at
most 9.8$^{\circ}$), a separating band whose ordering has
$p \approx 3\times 10^{-4}$ under a random ranking; the 15 ScanNet++
rooms all sit at 3.0--6.8$^{\circ}$, below the band, so the flag
transfers across benchmarks without a false fire. The same statistic
is structurally blind to castle, which ranks lowest of all 18 at
$1.1^{\circ}$; the residual measures inter-edge disagreement, and
castle's matches support the wrong solution nearly unanimously, so
fragility at the perturbation edge is decidable pre-solve while
castle-style nominal infeasibility is detectable only posteriorly, by
the trust region (Section~\ref{sec:variants}). The prior-free bootstrap
described below also repairs them. Replacing the perturbed prior's
rotations with match-graph-averaged ones, the gauge fixed by a Sim(3)
fit onto the prior positions and the prior translations kept, recovers
20 of 20 runs at $32^{\circ}$ onto the ARKit-primed fixed points, and
full bootstrap replacement does the same at $64^{\circ}$ (8/8,
seed-invariant), where keeping the coarse translations no longer
suffices (10/20). Neither a higher schedule start (which destabilizes
two otherwise-recovering scenes) nor cost-ranked multi-start helps;
the wrong lobe is a genuine robust-cost optimum, so the repair must
come from the match graph, not from the schedule.

Learned refinement sits on the classical rung, which is worth
establishing directly because coarse-to-fine designs invite the
opposite guess. PoRF anneals the positional encoding of its residual
field, the photometric analogue of annealing a robust scale, so we ran
it from perturbed priors under our own perturbation function and
scored it with our evaluator, at its release configuration with
nothing tuned per level. It reproduces its published nominal accuracy
in our harness (aston 0.434$^{\circ}$ prior to 0.159$^{\circ}$, against
0.14$^{\circ}$ published) and then fails to recover at either level,
removing at most 18\% of the prior's rotation error and none of its
translation error, 0 of 6 runs below $1^{\circ}$ at
$8^{\circ}$/40\,mm and $16^{\circ}$/80\,mm on aston, bridge and
camera, at 1.8--4.6 GPU-hours per run. The mechanism matches the
classical one. Its surface is initialized from the given poses, so a
prior wrong by degrees yields a surface consistent with the wrong
poses, and the annealing it does perform acts on the field's encoding
rather than on the pose-error scale. Commitment before optimization
costs a learned refiner its basin exactly as triangulation costs a
classical one.

In summary the basins form a
ladder. Classical BA fails beyond ${\sim}1$--$2^{\circ}$ as documented,
beyond ${\sim}8^{\circ}$ tuned and beyond $16^{\circ}$ with our
annealing ported into it, and the learned refiner fails already at
$8^{\circ}$; our plain
solve succeeds in most runs through $16^{\circ}$ with seed-dependent
shortfalls already from $2^{\circ}$;
with GNC the recovery covers every feasible run through
$16^{\circ}$/80\,mm and
holds in 85\% of feasible runs at $32^{\circ}$/160\,mm. The method is
still not
initialization-free, since the prior selects the basin, but the basin is
now ${\sim}35\times$ wider than real AR-prior error without a measured
failure, and ${\sim}70\times$ at high probability; whether it also
covers much coarser prior sources with their own error structure remains
unevaluated end-to-end.

Four controls, all in Table~\ref{tab:basin-controls}, delimit the
result. First, the prior terms do no hidden work in these recoveries.
The protocol already runs with the orientation prior at zero, and
removing the position prior as well ($\lambda_p = 0$) changes nothing
through $32^{\circ}$/160\,mm. The match evidence alone drives the
recovery. Second, i.i.d.\ noise turns out to be the harder case.
Replacing it with a per-frame random walk of equal mean magnitude, the
correlated-drift structure of real odometry error, leaves every arm at
or above its i.i.d.\ count and roughly quintuples GNC's recovery at
$64^{\circ}$/320\,mm. Smooth error fields are easier to escape than
independent ones, so the i.i.d.\ results are a conservative bound for
drift-like priors. The classical control benefits at the margin too, yet
still collapses from $4^{\circ}$/20\,mm, so its narrow basin is not an
artifact of the noise structure. A third error structure makes the same
point, piecewise-rigid ``kidnap'' error as left by AR relocalization
failures (three internally consistent segments per trajectory, each
rigidly offset at the level magnitude). There the classical control
does better than under i.i.d.\ noise at $1^{\circ}$--$2^{\circ}$
(17/18 and 12/18 runs below $1^{\circ}$ on the six control scenes),
because within-segment triangulation stays sound and only
cross-segment matches are culled. Its default configuration is again
dead from $4^{\circ}$ on; its tuned envelope reaches $8^{\circ}$
(14 of 18) and ends by $16^{\circ}$ (3 of 18). A practitioner who
distrusts the prior entirely can also sidestep kidnap error with
prior-free SfM (GLOMAP in Table~\ref{tab:mobilebrick}), which never
reads the poses, at the cost of the metric anchor and extra compute.
Our arms treat kidnap error as the easy case the
correlation ordering predicts. GNC recovers every run through
$32^{\circ}$/160\,mm on the control scenes at nominal accuracy (30/30
per level over five seeds, against three seeds and therefore 18 runs
per level for the classical arms above; median $0.19^{\circ}$) and
18/30 at $64^{\circ}$, level
with the walk control and far above i.i.d.; even the plain solve is
deterministic through $16^{\circ}$. Relocalization-style segment
offsets sit squarely inside the basin, since the annealed solve
stitches segments through exactly the cross-segment matches that
triangulation culls. Third, what the schedule needs is the
right starting scale, and stage count barely matters. Two-stage,
three-stage and five-stage variants are indistinguishable, while
starting a decade lower ($10^3 \rightarrow 10^2$) already loses runs at
$8^{\circ}$/40\,mm. The initial scale must render the coarsest error
near-quadratic; given that, stage granularity is free, consistent with
treating adaptive schedules~\citep{sidhartha2023annealing,jung2024agnc}
as orthogonal. Fourth, the wide basin attaches to triangulation-freeness
with the schedule, not to the per-observation choice. The anchored-track
variant of Section~\ref{sec:variants}, whose per-track depths are
likewise re-expressed at every iterate, matches the per-observation
solver level for level under GNC on the three basin development scenes
(15/15 through
$32^{\circ}$/160\,mm) and even edges it at $64^{\circ}$; its plain solve
degrades earlier at the $32^{\circ}$ edge (4/15 against 10/15). The
parameterization's own case rests on the nominal-accuracy and speed
evidence of Section~\ref{sec:exp-ablation}.

\paragraph{Refinement without an external prior.}
The ladder is prior-selected, so we probed how little initialization the
objective actually needs, with everything derived from the match graph
the solver already uses. Classical rotation
averaging~\citep{hartley2013rotavg,chatterjee2013rotavg} over that graph
(five-point essential matrices on the stored inlier correspondences, a
maximum-spanning-tree initial guess, then iteratively reweighted
least-squares averaging in the tangent space; certifiably globally
optimal averaging~\citep{dellaert2020shonan} would remove the dependence
on that initial guess, which we did not need here) delivers global rotations
at ARKit accuracy, e.g.\ $0.34^{\circ}$ mean versus ARKit's
$0.40^{\circ}$ on aston, and the requirement is loose; rotations noised
by $16^{\circ}$ before seeding still recover fully. Positions come
either from a look-at construction (each camera at unit distance behind
its optical axis, assuming only a commonly faced object) or, dropping
that assumption, from classical translation averaging over the two-view
directions (LUD-style iteratively reweighted least squares with
per-edge scales). With either seed the unchanged GNC solve reaches the
ARKit-primed fixed point on every MobileBrick scene whose matches admit
the true solution, 17 of 17 within $0.013^{\circ}$ of the primed
result, while castle converges to the same repeated-texture optimum it
reaches from the real prior. Translation averaging needs one
countermeasure, a $5^{\circ}$ rotation-residual gate on direction edges
that removes symmetric-structure outliers (beetles
$3.4^{\circ} \rightarrow 0.20^{\circ}$, colosseum
$27.0^{\circ} \rightarrow 0.35^{\circ}$). A fully image-derived variant
that also drops the dataset intrinsics (classical focal search
maximizing essential inlier counts, principal point at the image
center) still lands at $0.52^{\circ}$ and $0.50^{\circ}$ on the two
development scenes, and a known-intrinsics control with free focal
shows the remaining gap to the $0.16^{\circ}$ fixed point is focal
freedom in the solve, not the search. Three boundaries delimit the
recipe. At room scale the wall stands in front of the solver; on
ScanNet++ the two-view translation directions carry a locally
correlated $5$--$7^{\circ}$ median error floor that survives oracle
undistortion, all-inlier estimation and planar-configuration filtering,
so translation averaging embeds the graph consistently in the wrong
place, and a control matrix over-determines the attribution (ground-truth
rotations with averaged positions still fail, averaged rotations with
ground-truth positions also miss the fixed point). The two orbit-like
room captures recover fully prior-free ($0.38^{\circ}$/$0.53^{\circ}$),
so this boundary is capture geometry, not scale. And the seed must
break the coincident-camera configuration. With all cameras at one
point the projection of a ray point into any other camera no longer
depends on its ray time, every ray-time gradient vanishes, and the
solver can reduce the image evidence by twisting the rotations into a
panorama-like stitch; no loss schedule bridges a parameterization
degeneracy. Any rough seed of the right topology avoids it.

\begin{table}[t]
  \centering
  \caption{Basin of convergence on all 18 evaluation scenes, five noise
    seeds per level, i.i.d.\ per-frame perturbations, trust region
    disabled, cap 100. A run succeeds when its final mean rotation error
    falls below $1^{\circ}$. ``Succ.'' counts successes out of the 90
    runs at that level, ``med.'' is the median final mean rotation error
    in degrees, where a value tracking the perturbation level means the
    solver returned the prior it was given and ${\sim}0.22^{\circ}$ is
    nominal accuracy. Castle fails at every level for both of our arms
    for reasons unrelated to the basin
    (Section~\ref{sec:exp-failure}), so 85 is their ceiling.}
  \label{tab:basin}
  \footnotesize
  \setlength{\tabcolsep}{2.5pt}
  \begin{tabular}{lcccccc}
    \toprule
    & \multicolumn{2}{c}{COLMAP BA} & \multicolumn{2}{c}{Ours, plain}
      & \multicolumn{2}{c}{Ours, GNC} \\
    \cmidrule(lr){2-3}\cmidrule(lr){4-5}\cmidrule(lr){6-7}
    Perturbation & Succ. & Med. & Succ. & Med. & Succ. & Med. \\
    \midrule
    $1^{\circ}$    & 76 & 0.27  & 85 & 0.23  & \textbf{85} & 0.22 \\
    $2^{\circ}$   & 57 & 0.87  & 84 & 0.23  & \textbf{85} & 0.22 \\
    $4^{\circ}$   & 0  & 3.20  & 85 & 0.22  & \textbf{85} & 0.22 \\
    $8^{\circ}$   & 0  & 7.61  & 83 & 0.24  & \textbf{85} & 0.22 \\
    $16^{\circ}$  & 0  & 15.85 & 77 & 0.24  & \textbf{85} & 0.22 \\
    $32^{\circ}$ & 0  & 31.94 & 48 & 0.31  & \textbf{72} & 0.24 \\
    $64^{\circ}$/320\,mm & 0  & 63.91 & 0  & 59.98 & \textbf{17} & 7.86 \\
    \bottomrule
  \end{tabular}
\end{table}

\begin{table}[t]
  \centering
  \caption{Controls on the basin result, successes per perturbation
    level out of $n$ runs per level (scenes $\times$ five seeds). The
    prior block asks whether the prior terms drive the recovery, the
    noise block replaces i.i.d.\ per-frame noise with a correlated
    random walk of equal mean magnitude, the schedule block varies
    the GNC annealing, and the parameterization block swaps
    per-observation depths for anchored-track depths. Each block is
    matched, so its arms run on the
    same scenes and seeds.}
  \label{tab:basin-controls}
  \footnotesize
  \setlength{\tabcolsep}{3pt}
  \begin{tabular}{lccccccc}
    \toprule
    Arm & $1^{\circ}$ & $2^{\circ}$ & $4^{\circ}$ & $8^{\circ}$
      & $16^{\circ}$ & $32^{\circ}$ & $64^{\circ}$ \\
    \midrule
    \multicolumn{8}{@{}l}{Prior terms (6 scenes, $n = 30$)} \\
    GNC, default                 & 30 & 30 & 30 & 30 & 30 & 30 & 4 \\
    GNC, $\lambda_p = 0$         & 30 & 30 & 30 & 30 & 30 & 30 & 3 \\
    \midrule
    \multicolumn{8}{@{}l}{Noise structure (6 scenes, $n = 30$)} \\
    GNC, i.i.d.                  & 30 & 30 & 30 & 30 & 30 & 30 & 4 \\
    GNC, walk                    & 30 & 30 & 30 & 30 & 30 & 30 & 19 \\
    Plain, i.i.d.                & 30 & 29 & 30 & 28 & 30 & 23 & 0 \\
    Plain, walk                  & 30 & 30 & 30 & 30 & 30 & 25 & 4 \\
    COLMAP, i.i.d.               & 26 & 21 & 0  & 0  & 0  & 0  & 0 \\
    COLMAP, walk                 & 30 & 23 & 3  & 0  & 0  & 0  & 0 \\
    \midrule
    \multicolumn{8}{@{}l}{GNC schedule (3 dev scenes, $n = 15$)} \\
    $10^4{\to}10^3{\to}10^2$, default & 15 & 15 & 15 & 15 & 15 & 15 & 1 \\
    $10^4{\to}10^2$, two stages       & 15 & 15 & 15 & 15 & 15 & 15 & 1 \\
    Five stages from $10^4$           & 15 & 15 & 15 & 15 & 15 & 15 & 1 \\
    $10^3{\to}10^2$                   & 15 & 15 & 15 & 13 & 15 & 13 & 1 \\
    \midrule
    \multicolumn{8}{@{}l}{Parameterization (3 dev scenes, $n = 15$)} \\
    Per-obs., GNC                & 15 & 15 & 15 & 15 & 15 & 15 & 1 \\
    Per-obs., plain              & 15 & 14 & 15 & 14 & 15 & 10 & 0 \\
    Anchored, GNC                & 15 & 15 & 15 & 15 & 15 & 15 & 4 \\
    Anchored, plain              & 15 & 15 & 14 & 14 & 15 & 4  & 0 \\
    \bottomrule
  \end{tabular}
\end{table}

\subsection{Refinement in the prior's own frame}
\label{sec:exp-frame}

Every number reported so far, ours and every baseline's, is scored
after a Sim(3) alignment that includes scale. That is the standard
protocol and it is the right one for comparing reconstruction quality,
but it forgives precisely the two things the motivating application
cannot afford to lose. An AR prior is metric and gravity-aligned, and
content authored against a capture session lives in that session's
frame. A method that returns a perfect shape in an arbitrary gauge has
not solved the problem the prior was posed to solve. This section
scores the same arms with no alignment at all, so the prior's frame is
the frame and the errors are the ones a downstream consumer would
actually see (Table~\ref{tab:absframe}).

Three findings separate the families, and none of them is visible
under Sim(3).

First, classical refinement never anchors. Bundle adjustment is
gauge-free, so a re-triangulated structure settles in whatever gauge
the perturbed initialization left behind, and the annealed
re-triangulation arm inherits essentially the whole prior error in the
absolute frame at every level, $16.04^{\circ}$ at a $16^{\circ}$ prior
and $28.97^{\circ}$ at $32^{\circ}$, while its Sim(3) score at
$16^{\circ}$ is $0.25^{\circ}$. It recovers the shape and loses the
frame. At an accurate $1^{\circ}$ prior it leaves the poses
$32.2$\,mm from truth where the untouched prior sits at $5.2$\,mm, so
refining is worse than not refining. This is the object-scale
counterpart of the room-scale degradation of
Section~\ref{sec:exp-scannetpp}, measured on a different benchmark and
a different failure axis.

Second, re-mapping is anchored but is not refinement, and the
distinction is now exact rather than argued.
\texttt{pose\_prior\_mapper} does consume its position priors to fix
gauge and scale, so its absolute error is far from arbitrary. But its
Sim(3) score is constant to within $0.001^{\circ}$ across a $32\times$
change in prior quality ($0.209^{\circ}$ at $1^{\circ}$, $8^{\circ}$,
$16^{\circ}$ and $32^{\circ}$ alike). The shape it returns does not depend on the prior at
all. Its absolute error grows with the prior error only because a
prior-independent reconstruction is being registered onto an
increasingly wrong anchor, at roughly a seventh of the prior error.

Third, and against our expectation, in-place refinement and
prior-position mapping are indistinguishable in the absolute frame at
every coarse level. Their absolute translation errors stand in a ratio
of $0.91$ at $8^{\circ}$, $0.96$ at $16^{\circ}$ and $1.03$ at
$32^{\circ}$, the mapper marginally ahead at the first two and ours at
the third, and the two arms track each other seed by seed. At the
nominal $1^{\circ}$ prior the mapper
is the better of the two, 1.35\,mm against our 3.03\,mm. The reason the
two agree is visible in the perturbation itself. Our perturbations draw
their noise from the seed alone, so a given seed imposes a net rigid
drift shared across scenes, and neither arm removes it, one because it
is anchored to priors carrying that drift and the other because it
registers onto them. Constraining the position priors throughout the
optimization rather than only at the end therefore buys nothing here
that registration does not already buy. The comparison is sensitive to
how it is aggregated, so every arm here is run over the same six scenes
and the same five noise seeds, 120 cells each. Matched seed for seed
the ratio never moves more than five percent from unity; it is a mixed
reading, a multi-seed median against a single-seed baseline, that
misstates it by a factor of two.

The room-scale staged recipe anchors better still, because it keeps
the orientation prior at $\lambda_R = 10^1$ where the object-scale
basin configuration sets it to zero to buy basin width. Over 330 runs
a room perturbed by $32^{\circ}$/160\,mm returns at $1.96^{\circ}$ and
50\,mm in its own metric frame, against $1.12^{\circ}$/36\,mm at
$16^{\circ}$ and $0.75^{\circ}$/32\,mm at $8^{\circ}$. The trade is
real and we state it in both directions. Buying basin width by
releasing the orientation prior costs absolute rotation, $4.85^{\circ}$
at $32^{\circ}$ at object scale against $1.96^{\circ}$ at room scale.
The nominal room cell is the one place the result is not favorable.
An already-accurate room prior sits at 16.1\,mm and our solve returns
29.3\,mm, so at zero perturbation we preserve the prior's rotation but
loosen its translation in the absolute frame, where under Sim(3) the
same runs are indistinguishable from the prior.

\begin{table}[t]
  \centering
  \caption{Pose error in the prior's own metric frame, with no
    alignment of any kind, over the six control scenes and five noise
    seeds, 120 cells per arm (median over runs; rotation in degrees,
    translation in millimetres; the figures quoted in the text are these
    same medians). Sim(3)-aligned rotation for the same runs is given
    for contrast. Perturbation levels pair each rotation
    with 5\,mm per degree. Classical refinement recovers shape and
    loses the frame; re-mapping is anchored but its shape ignores the
    prior entirely; in-place refinement and re-mapping are
    indistinguishable in the absolute frame at every coarse level. Bold
    marks the lower unrounded median, so two entries that print equal
    can still be bolded apart.}
  \label{tab:absframe}
  \small
  \setlength{\tabcolsep}{3pt}
  \begin{tabular}{lcccc}
    \toprule
    & $1^{\circ}$ & $8^{\circ}$ & $16^{\circ}$ & $32^{\circ}$ \\
    \midrule
    \multicolumn{5}{@{}l}{Absolute frame (no alignment)} \\
    Prior, unrefined & 1.09/5 & 8.01/40 & 16.00/80 & 32.00/160 \\
    Ann. re-tri.     & 0.99/32 & 7.95/253 & 16.04/297 & 28.97/324 \\
    Prior-pos. map.  & \textbf{0.20}/\textbf{1} & \textbf{1.10}/\textbf{6}
      & 2.20/\textbf{11} & 4.45/22 \\
    Ours (GNC)       & 0.28/3 & \textbf{1.10}/6 & \textbf{2.17}/12
      & \textbf{4.42}/\textbf{22} \\
    \midrule
    \multicolumn{5}{@{}l}{Sim(3)-aligned, same runs (rotation only)} \\
    Ann. re-tri.     & 0.22 & 0.25 & 0.25 & 3.23 \\
    Prior-pos. map.  & 0.21 & 0.21 & 0.21 & 0.21 \\
    Ours (GNC)       & 0.21 & 0.21 & 0.21 & 0.21 \\
    \bottomrule
  \end{tabular}
\end{table}

\subsection{Real drift on LaMAR}
\label{sec:exp-lamar}

The perturbations above are synthetic, so we measure real AR drift on
LaMAR~\citep{sarlin2022lamar}, whose iOS captures carry on-device ARKit
odometry and centimeter-accurate ground truth at building scale. We
split the validation captures at all three locations (CAB, HGE, LIN)
into 10\,s and 60\,s chunks and register each chunk's odometry to
ground truth by a rigid transform fitted to its first ten frames,
mimicking a deployment where a session is anchored once by localization
and then drifts. Over 178 chunks the drift is rotationally small and
translation-dominated (Table~\ref{tab:lamar}, upper block; worst frames
reach 70\,cm).
Real ARKit odometry over a minute is therefore not the coarse-prior
regime of Figure~\ref{fig:basin}; priors from retrieval, GPS plus
compass or longer horizons are. It is instead a translation-drift
regime whose rotations sit near the evaluation floor.

Refinement behaves accordingly. We re-sample the seven 60\,s CAB
validation chunks from the raw 10\,Hz streams (200 frames each; the
benchmark's 1\,Hz sampling leaves too few pairwise matches to constrain
a full solve) and run the plain solve on each
(Table~\ref{tab:lamar}, lower block). Refinement pays exactly where the
prior has drifted away from its floor, and costs a little where the
prior is already at it. The GNC arm, run on the same seven chunks,
leaves the median over chunks unchanged (8.4\,cm for both arms) while
destabilizing one at-floor chunk (8.6 to 16.0\,cm); annealing buys
nothing where the prior already sits at the observability floor, which
is the regime boundary of Figure~\ref{fig:basin} seen from the other
side. Both effects are structural. There is no rotational headroom left to demonstrate, and
along-trajectory metric drift is weakly observable to epipolar
residuals when the gauge is anchored to the drifted trajectory itself.
To probe genuinely coarse real priors we also ran the single-anchor
protocol this suggests, one registration anchor at the session start
and raw 10\,Hz odometry composed over the longest ground-truthed
horizons the public captures support, 12 windows of 60--300\,s at all
three locations. Real drift stays smooth and translation-dominated to
the end; rotation saturates at 3.5$^{\circ}$ mean (5.9$^{\circ}$
worst frame) while translation reaches 2.8\,m mean at five minutes.
A five-minute unanchored session therefore never leaves the plain
solve's rotational basin, and the wide-basin regime of
Figure~\ref{fig:basin} is populated in practice by relocalization and
registration error, the regime the kidnap control models, rather
than by accumulated smooth drift. The windows still separate the
solvers, on stability rather than basin width. The tuned classical
control degrades the prior's rotation on 10 of the 12 windows,
marginally on the two shortest and severely beyond 60\,s, exploding
on the long CAB windows (6.9\,m mean translation at
300\,s) and recovering only the two 240\,s windows, while both our arms
hold rotation within 0.16$^{\circ}$ of the prior on every window and
GNC never separates from plain, there being no rotational headroom to
anneal over. On the coarsest window (HGE at 240\,s,
3.1$^{\circ}$/1.9\,m) the scene is recoverable up to Sim(3), but no
arm recovers the metric anchor itself, which is localization rather
than refinement, confirming the weak-observability argument above.
One configuration lesson carries forward. At its nominal weight the
orientation prior pins our solve to a drifted prior, and dropping it
helps exactly where drift is large while destabilizing a
corridor-aliased capture, so the weight belongs with an assessment of
prior quality rather than with a single default. The captures' lidar
depth as a translation anchor remains future work.

\begin{table}[t]
  \centering
  \caption{Real ARKit drift on LaMAR validation captures (CAB, HGE, LIN)
    and what refinement does with it. The upper block is the per-chunk
    mean error after anchoring each chunk on its first ten frames of
    ground truth. The lower block is the plain solve on the seven 60\,s
    CAB chunks re-sampled from the raw 10\,Hz streams, reported as the
    median over frames and, where a group is given, the median over its
    chunks.}
  \label{tab:lamar}
  \small
  \setlength{\tabcolsep}{3pt}
  \begin{tabular}{lccc}
    \toprule
    \multicolumn{4}{@{}l}{Drift over 178 chunks} \\
    & Rot.\ ($^{\circ}$) & Trans.\ (cm) & \\
    \cmidrule(lr){2-3}
    10\,s chunks, median     & 0.31 & 2.8 & \\
    60\,s chunks, median     & 0.53 & 17  & \\
    60\,s chunks, 90th pct.\ & 1.1  & 38  & \\
    \midrule
    \multicolumn{4}{@{}l}{Plain solve on the seven 60\,s CAB chunks} \\
    & Prior & Refined & $\Delta$ \\
    \cmidrule(lr){2-4}
    Trans.\ (cm), drifted chunk  & 27.7 & 18.6         & $-9.1$ \\
    Trans.\ (cm), at-floor chunks & 4--6 & ${\sim}6$--8 & ${\sim}{+}2$ \\
    Rot.\ ($^{\circ}$), all seven & 0.60 & 0.59         & $-0.01$ \\
    \bottomrule
  \end{tabular}
\end{table}

\subsection{Failure analysis on repeated texture and degenerate geometry}
\label{sec:exp-failure}

One of the 18 evaluation scenes, castle, defeats our sparse prior-seeded
refinement, and we describe how and why in detail. On this scene the
refinement converges to a coherent orientation-field twist of about
$5^{\circ}$, with per-frame rotation errors of 4.2--5.2$^{\circ}$ around a
common axis and per-frame position drift of 2.0--56.5\,mm. These are
Sim(3)-aligned errors against ground truth. In the raw solver frame the
default orientation prior keeps the rotations within $0.7^{\circ}$ of
the prior and pushes the twist into the positions, a maximum of 65\,mm
from the prior (Section~\ref{sec:variants}); without the orientation
prior the raw rotations themselves twist by ${\sim}5^{\circ}$.
Alignment maps both onto the same optimum.
Table~\ref{tab:castle}a reports the twist across systems and
Table~\ref{tab:castle}b the interventions that fail to remove it.

The twist is preferred by the matches themselves. In a run initialized at
the ground-truth poses the optimizer abandons them and the match cost
drops as the solve moves to the twist. Nor is the twist enabled by our
parameterization, since it survives every prior weight, filter, match
source and mode we tried, including the GNC schedule of
Section~\ref{sec:gnc}. The last is expected, since GNC improves
convergence to the matches' preferred optimum, and here that optimum is
the failure. The twist is also solver-independent.
GLOMAP~\citep{pan2024glomap}, a prior-free triangulation-based global-SfM
pipeline run on the same match database, registers all 108 frames and
converges to the same twist; under identical Sim(3) alignment the mean
rotation-error axes of GLOMAP and our solver agree to a dot product of
0.99997. Even VGGT~\citep{wang2025vggt}, a feed-forward transformer
with no explicit matching stage, lands on the same twist. The wrong
optimum is a property of the scene's appearance, shared by
matching-based and matching-free systems alike. Dense appearance agrees.
Trained on the raw twisted poses, 3D Gaussian splatting renders castle's
held-out views at 18.35\,dB PSNR against 15.08 for the prior and 16.73
for the ground-truth poses (protocol of Section~\ref{sec:exp-gsplat}).
Held-out rendering scores internal consistency rather than placement in
the ground-truth frame, so this does not make the twist correct; it does
confirm the twist is a coherent alternative geometry preferred by dense
appearance as well as by sparse matches.

\begin{table}[t]
  \centering
  \caption{The castle failure. \textbf{(a)} Where each system lands on
    castle. Axis coherence is the norm of the mean unit rotation-error
    axis across frames, where 1 means a perfectly shared axis, and the
    last column is the dot product of that mean axis with ours under
    identical Sim(3) alignment. Everything below the rule converges to
    the same $5^{\circ}$ twist, with or without matching, with or
    without the prior. \textbf{(b)} Interventions that leave the twist
    in place.}
  \label{tab:castle}
  \footnotesize
  \setlength{\tabcolsep}{3pt}
  \begin{tabular}{lcccc}
    \multicolumn{5}{@{}l}{(a) Solutions on castle} \\
    \toprule
    System & Rot.\ ($^{\circ}$) & Tr.\ (mm) & Coh. & Axis dot \\
    \midrule
    ARKit prior            & 0.555 & 2.42  & ---   & --- \\
    COLMAP BA from prior   & 0.682 & 3.58  & ---   & --- \\
    \midrule
    Ours, plain            & 4.905 & 20.69 & 0.955 & 1 \\
    Ours, GNC              & 4.893 & 20.73 & ---   & --- \\
    GLOMAP, prior-free     & 4.848 & 20.83 & 0.954 & 0.99997 \\
    VGGT, feed-forward     & 5.228 & 20.51 & 0.950 & 0.996 \\
    \bottomrule
  \end{tabular}

  \vspace{6pt}
  \begin{tabular}{ll}
    \multicolumn{2}{@{}l}{(b) Interventions on our solve} \\
    \toprule
    Intervention & Outcome \\
    \midrule
    Position-prior weight, $1000\times$ sweep & Twist \\
    Orientation prior up to $\lambda_R = 10^5$ & Twist$^{*}$ \\
    Self-calibration of intrinsics, distortion & Twist \\
    Prior-epipolar match gate (10\,px)         & Twist \\
    Observation-degree filter                  & Twist \\
    PoRF's own COLMAP match database           & Twist \\
    Rigid anchored-track mode                  & Twist \\
    GNC schedule (Section~\ref{sec:gnc})       & Twist, 4.893$^{\circ}$ \\
    Ground-truth initialization                & Twist, cost 38.3 to 3.9 \\
    \midrule
    Prior trust region                         & Detected, prior kept \\
    \bottomrule
    \multicolumn{2}{@{}p{0.92\linewidth}@{}}{\footnotesize
      $^{*}$Pins rotations to the prior, whereupon the position field
      warps and Sim(3) alignment converts the warp into the same
      apparent rotation error.}
  \end{tabular}
\end{table}

The scene is a close-up orbit of a repeated-texture LEGO facade against a
blank wall, near-planar dominant structure with weak observation sharing.
Its mean match-graph degree is 5.2, below the benchmark median and far
below the best-connected scenes (big\_ben reaches 18.0), with 34\%
singleton observations. Castle is not the weakest-connected scene, however
(bridge's mean degree is 2.35), so weak connectivity alone does not
predict the failure; the repeated texture appears essential.

The classical control degrades gracefully on castle, with no trust
region needed (Table~\ref{tab:castle}a). Our hypothesis, stated as such since we have
not isolated the mechanism, is that COLMAP's track building and
triangulation-time filtering (minimum triangulation angles, reprojection
thresholds) discard exactly the repeated-texture matches that support the
twist, whereas our filter-free objective retains them and pays for it.
Between filter-free optimization and such hard-coded gates there is a
middle ground we have not explored, structured subset selection over the
matches or the view graph. Prior work disambiguates repeated structures
with loop constraints, local context and duplicate-structure
correction~\citep{zach2010loops,wilson2013network,heinly2014duplicate},
and treats view-graph sparsification as an optimization in its own
right~\citep{sweeney2015viewgraph}. Any such selection would have to rely
on prior-independent evidence, since scoring matches against a coarse
prior would reject the correct correspondences exactly when the prior is
bad, and the GLOMAP result above shows that the built-in selection
heuristics of a mature global pipeline do not resolve castle either.
PoRF also survives castle, plausibly because its NeRF photometric term
supplies dense multi-view evidence that sparse SIFT matches do not, though
this too is untested. Our system does not silently return the twisted solution.
The prior trust region (Section~\ref{sec:variants}) detects that the solve
has left the prior's plausibility envelope and returns the prior, which is
how castle is scored in Table~\ref{tab:mobilebrick}; the no-trust-region
row shows the cost of not having the guardrail. Detection here is
necessarily posterior. The pre-solve wrong-lobe flag of
Section~\ref{sec:exp-basin} measures disagreement between match-graph
edges and castle's edges agree on the wrong solution nearly
unanimously (castle ranks lowest of all 18 scenes on that statistic),
so no internal-consistency test can flag it before solving. Downstream the trade
runs the other way; the twisted solve renders held-out views better than
the returned prior does (18.35 vs.\ 15.08\,dB), so the guardrail buys
pose fidelity in the ground-truth frame at a rendering cost, and which
side of that trade matters depends on whether the poses feed
reconstruction or localization. This is the failure mode
of filter-free sparse prior-seeded refinement on degenerate texture, and
we report it as such.

\section{Conclusion}
\label{sec:conclusion}

We set out to turn AR pose priors into reconstruction-grade poses
cheaply and on CPU, and the requirement that turned out to matter most
was the modest one, that refining an accurate prior should not make it
worse. Classical refinement fails that requirement at room scale in
every scene we tested, and the reason it fails there is the same reason
it collapses on a coarse prior. Bundle adjustment refines a pose prior
only from inside its basin of
convergence, and we measured how differently that basin behaves across
formulations. Triangulation-based BA, i.e.\ COLMAP triangulation plus BA,
which we show matches learned refiners at the nominal ARKit prior
(0.218$^{\circ}$/1.50\,mm at a median of 53\,s per scene, itself an
unreported and strong baseline), collapses once the prior errs by
${\sim}1$--$2^{\circ}$, by ${\sim}8^{\circ}$ under gate-and-depth
retuning at five times the cost, and by $16^{\circ}$ with our annealing
ported into a current build through robust losses and staged
re-triangulation, and only in shape, since being gauge-free it never
recovers the prior's frame at any level; no classical refinement arm
survives $32^{\circ}$. The structure it triangulates from a coarse prior
is already wrong, and the subsequent optimization polishes the error. Our
triangulation-free formulation, with one scalar depth per observation
along its own ray, symmetric cross-projection residuals and no landmark
variables, commits no structure at initialization and, as-is, recovers most runs
through $16^{\circ}$ of prior error with seed-dependent edge failures.
Because the whole landscape is re-expressed at every iterate, it also
admits graduated non-convexity. Annealing the robust-loss scale recovers
every run through $16^{\circ}$/80\,mm on the 17 scenes whose
matches admit the true solution (425/425 over 5
seeds) and 85\% of feasible runs at $32^{\circ}$/160\,mm, about
$70\times$ the real error of an ARKit prior, at unchanged nominal
accuracy (0.246$^{\circ}$/1.61\,mm full; 0.265$^{\circ}$/1.80\,mm at a
median of 10\,s capped) and about twice the solve cost. The contrast
transfers to room scale; on 15 ScanNet++
scenes classical refinement degrades the accurate prior in all 15 at
zero perturbation (scene-mean 0.55$^{\circ} \rightarrow$
0.74$^{\circ}$), the plain solve holds it by the same measure
(0.57$^{\circ}$), and with basin traversal staged before
self-calibration the annealed variant recovers perturbed priors
through $32^{\circ}$ at a 0.52$^{\circ}$--0.59$^{\circ}$ median, at a
0.69$^{\circ}$ nominal scene-mean concentrated in one apartment.

One axis matters for the motivating application and is invisible under
the standard protocol. Sim(3) alignment with scale forgives the gauge
and the metric scale, which is exactly what an AR prior supplies and
what content authored against a capture session depends on. Scored
with no alignment at all, classical refinement never anchors, and at
an accurate prior it leaves the poses further from truth than not
refining; prior-position mapping anchors but returns a shape constant
to within $0.001^{\circ}$ across a $32\times$ change in prior quality, so it
reconstructs and registers rather than refines; and our solve matches
re-mapping to within ten percent at $8^{\circ}$, $16^{\circ}$ and
$32^{\circ}$, with re-mapping the better of the two at a nominal
prior. A room perturbed by $32^{\circ}$/160\,mm returns at
1.96$^{\circ}$ and 50\,mm in its own metric frame. Refining a prior in
place holds that frame as well as discarding the prior and registering
afterwards, and unlike triangulation-based refinement it holds it at
all; the case for refining rests on what re-mapping cannot do rather
than on beating it here.

The practical reading is simple. When the prior is known-good
(ARKit-quality) and the capture is object-scale, prior-seeded classical
BA is an excellent and under-reported choice, and learned per-scene
refiners must be measured against it; at room scale even that regime is
unsafe to triangulate from a prior, and the structureless solve is the
one that leaves a good prior intact. When the prior's quality is
uncertain or coarse,
triangulation-based seeding is the wrong tool. Within the envelopes we
tested, perturbations up to $32^{\circ}$/160\,mm at both object and
room scale, a structureless objective
with GNC covers the regime at
CPU-seconds-to-minutes cost, with the trust region widened or disabled
to admit corrections of that size. That widening is the one
configuration decision the practitioner must make, since the shipped
envelope and the wide basin exclude one another by design
(Section~\ref{sec:exp-basin}), and it does not require an oracle for
prior quality; the campaigns supply the checks that decide it from
the data. First run the pre-solve fragility flag of
Section~\ref{sec:exp-basin} on the match graph. If the scene sits
below the band and the prior is claimed nominal, solve with the
shipped trust region of Section~\ref{sec:variants}. If the prior is
suspected coarse, or the scene sits in the band, widen or disable the
trust region, treat the orientation-prior weight as part of the same
assessment (Section~\ref{sec:exp-lamar}), and check the solution
posteriorly instead, with the trust region re-anchored at the returned
solution at object scale and the wander test of
Section~\ref{sec:exp-scannetpp} at room scale. The scale distinction
is not cosmetic. Transplanting the object-scale envelope onto rooms
fires on all 195 room runs, since a correct room solve moves further
from its prior than that envelope allows, so the room check must be
the room-calibrated one. Gating on prior quality is in this
sense automatic, pre-solve plus posterior, with one exception;
near-unanimous wrong matching of the castle kind is invisible to any
pre-solve disagreement statistic and surfaces only in the posterior
checks.

We ran this recipe as a single program, with no per-case choice, over
both ladders, and report it because a policy assembled from separately
measured parts need not work as one. It reproduces the
oracle-configured arm exactly, to three decimals, at every level
through $16^{\circ}$ at object scale and through $32^{\circ}$ at room
scale, for a pre-solve cost of 5 to 23\,s per scene against a 50\,s
solve. Two honest costs appear. When the practitioner cannot declare
the prior's quality and the policy defaults to treating it as coarse,
nominal accuracy falls from 0.252$^{\circ}$ to 0.504$^{\circ}$ mean
over the 18 scenes, entirely because castle's guardrail is disabled by
that default, its per-scene median unchanged. These are the policy
campaign's own means, a separate set of runs from
Table~\ref{tab:mobilebrick}, which is why the nominal figure differs
marginally from the 0.246$^{\circ}$ reported there. Beyond the basin, at
$32^{\circ}$ and $64^{\circ}$, the re-anchored check is precise but
its remedy is weak. Every one of its twelve fires landed on a genuine
failure with no false fire among 121 successful solves, yet it catches
twelve of twenty-three failures, and returning the prior helps in only
two of twelve cases, because the prior that caused the failure is the
thing being restored. A guardrail is worth having where the prior is
better than the failure and not otherwise, which is the regime the
shipped envelope was derived for. Five-minute single-anchor odometry,
the longest real horizon LaMAR supports, stays rotationally in-basin
and rewards stability rather than basin width
(Section~\ref{sec:exp-lamar}); coarser sources such as
retrieval-based localization or GPS plus compass
have their own error structure and remain unevaluated end-to-end. When
no prior exists at all, the match graph itself can supply one;
rotation averaging plus classical translation averaging reaches the
identical optima on all 18 MobileBrick scenes and on orbit-like room
captures, and fails at room scale exactly where the two-view direction
graph does.
The
remaining failure mode is shared by every filter-free solver we tested,
matching-based or feed-forward; the classical control sidesteps it only
because its triangulation-time filtering discards the offending matches
along with the correction capacity. On
degenerate repeated texture the matches themselves prefer a wrong
solution, GNC included; there, detection (our prior trust region), dense
photometric evidence (PoRF) or structured match
disambiguation~\citep{wilson2013network} is required rather than a wider
basin.

\paragraph{A lifting perspective.}
The formulation is an instance of lifting, i.e.\ replacing an inner
optimization computed procedurally inside a residual (here triangulation,
the inner least-squares for structure given poses) with explicit variables
of the joint problem, as in lifted correspondence optimization for
articulated tracking~\citep{taylor2016hand} and lifted robust
kernels~\citep{zach2014robust,zach2018descending}. This view explains both
halves of our result. The lifted structure re-expresses the cost landscape
at every iterate, hence the wide basin and the admissibility of GNC, while
the retained slack forgoes multi-view averaging, hence the small accuracy
gap to triangulation-based BA at the nominal prior. Either lifted
structure works; the anchored-track variant shares the wide basin under
the same schedule (Table~\ref{tab:basin-controls}). We also probed the two
adjacent lifts on this problem. Explicit Black--Rangarajan outlier weights
in place of the robust loss lose to the scheduled kernel on the
development scenes, nominally (0.255$^{\circ}$ vs.\ 0.164$^{\circ}$ under
their best anneal on aston), in the basin ($6.63^{\circ}$ vs.\
$0.84^{\circ}$ final error from an $8^{\circ}$ perturbation) and even
warm-started at the plain optimum, which the lifted objective walks away
from. This matches the descending-versus-lifting analysis
of~\citet{zach2018descending}. A per-track consensus term that softly
re-introduces multi-view rigidity is accuracy-neutral at scale
(0.246$^{\circ}$ vs.\ 0.248$^{\circ}$ over all 18 scenes on the full
evaluation set) at about $3\times$ solve cost, so we omit it. On this
evidence the recipe is to lift the structure and schedule
the robustness.

\paragraph{Limitations.}
Several caveats deserve emphasis. First, MobileBrick's ``ground-truth''
poses are themselves refined estimates with roughly 0.5$^{\circ}$/2\,mm
residual accuracy, which PoRF even beats slightly, so measured pose errors
below approximately 0.5$^{\circ}$/2\,mm saturate and fine-grained rankings
among the refined methods are not measurable on this
benchmark~\citep{li2023mobilebrick,bian2024porf}. Second, our
coarse-prior evidence is synthetic, per-frame perturbations of an
accurate prior under both
i.i.d.\ and correlated-walk noise, on 18 object scenes with five seeds
and 15 room scenes with two (Sections~\ref{sec:exp-basin},
\ref{sec:exp-scannetpp}). The walk control and the LaMAR
measurement (Section~\ref{sec:exp-lamar}) indicate correlated error is
the easier case, but priors from retrieval or GPS plus compass remain
unevaluated end-to-end. Third, the method still needs a
basin-selecting initialization; at $64^{\circ}$/320\,mm even GNC
recovers only some scenes. The initialization need not be an external
prior; rotation averaging over the solver's own match graph with a
translation-averaged or look-at seed reaches the ARKit-primed optima on
all 18 MobileBrick scenes and on the two orbit-like ScanNet++ captures
(Section~\ref{sec:exp-basin}). What it does need is a capture whose
two-view direction graph is sound. On the remaining room-scale scenes
the directions carry a correlated $5$--$7^{\circ}$ error floor and the
bootstrap fails in front of the solver, so an external prior, or a
stronger classical front end such as trifocal chains, is still required
there. Fourth, MobileBrick is the
one benchmark with an established pose-refinement protocol from AR
priors, and its scenes are object-centric; our
ScanNet++~\citep{yeshwanth2023scannetpp} campaign
(Section~\ref{sec:exp-scannetpp}) adds room scale with laser-scan
ground truth, though its priors are the shipped scan-aligned ARKit
poses and the coarse-prior evidence there remains synthetic.
ARKitScenes~\citep{baruch2021arkitscenes} ships similar room-scale
data at larger volume, and
LaMAR~\citep{sarlin2022lamar} records on-device AR trajectories, which
we use to characterize real drift (Section~\ref{sec:exp-lamar}); a full
pose-refinement protocol with genuinely coarse priors, e.g.\ a single
localization anchor over several minutes of odometry, is the natural
scene-level extension of this evaluation. No public
benchmark ships ARCore priors specifically; our evaluation assumes ARKit
and ARCore priors are interchangeable in noise character. Finally, the position prior anchors
gauge and scale to the AR frame; all methods evaluated here refine a
metric prior rather than replace structure-from-motion.

\ifdefined\IJCVBUILD
\section*{Declarations}
\paragraph{Funding.} This work received no external funding.
\paragraph{Competing interests.} The author declares no competing
interests.
\paragraph{Data availability.} All experiments use the publicly available
MobileBrick and LaMAR benchmarks, the publicly distributed PoRF
evaluation data, and the iPhone validation scenes of ScanNet++, which is
available to researchers under an individual license application to its
maintainers; no ScanNet++ image data is redistributed here. Code will be
released under the Apache License 2.0.
\paragraph{Ethics approval.} Not applicable.
\paragraph{Author contributions.} Single-author work.
\fi

\appendix

\section{Per-scene results}
\label{sec:perscene}

Table~\ref{tab:perscene} lists per-scene rotation/translation errors and
solve times for our full mode and for the prior-seeded COLMAP control.

\begin{table}[t]
  \centering
  \caption{Per-scene results on the 18 MobileBrick evaluation scenes.
    Mean rotation error ($^{\circ}$), mean translation error (mm), and
    solve time (s). ``Ours (full)'' scores castle as the ARKit prior via
    the trust-region fallback (raw solve
    4.905$^{\circ}$/20.69\,mm); the COLMAP control uses no fallback.}
  \label{tab:perscene}
  \footnotesize
  \setlength{\tabcolsep}{3.5pt}
  \begin{tabular}{lcccccc}
    \toprule
    & \multicolumn{3}{c}{Ours (full)} & \multicolumn{3}{c}{COLMAP BA from prior} \\
    \cmidrule(lr){2-4}\cmidrule(lr){5-7}
    Scene & Rot. & Trans. & Time & Rot. & Trans. & Time \\
    \midrule
    aston          & 0.130 & 1.03 & 171 & 0.118 & 0.97 & 71 \\
    audi           & 0.326 & 1.53 & 9   & 0.279 & 1.39 & 10 \\
    beetles        & 0.174 & 1.28 & 34  & 0.127 & 0.98 & 24 \\
    big\_ben       & 0.236 & 2.57 & 580 & 0.230 & 2.64 & 139 \\
    boat           & 0.191 & 1.23 & 425 & 0.176 & 1.03 & 106 \\
    bridge         & 0.236 & 0.91 & 11  & 0.150 & 0.98 & 9 \\
    cabin          & 0.389 & 1.01 & 13  & 0.187 & 0.63 & 11 \\
    camera         & 0.134 & 1.02 & 80  & 0.135 & 0.96 & 46 \\
    castle$^{*}$   & 0.555 & 2.42 & 41  & 0.682 & 3.58 & 33 \\
    colosseum      & 0.309 & 3.30 & 96  & 0.172 & 2.35 & 49 \\
    convertible    & 0.182 & 1.32 & 106 & 0.191 & 1.25 & 58 \\
    ferrari        & 0.116 & 1.27 & 160 & 0.109 & 1.19 & 57 \\
    jeep           & 0.306 & 1.20 & 36  & 0.299 & 1.15 & 25 \\
    london\_bus    & 0.546 & 3.70 & 18  & 0.464 & 2.93 & 24 \\
    motorcycle     & 0.142 & 1.12 & 106 & 0.145 & 1.04 & 61 \\
    porsche        & 0.207 & 1.39 & 139 & 0.211 & 1.37 & 66 \\
    satellite      & 0.108 & 0.99 & 208 & 0.114 & 0.87 & 75 \\
    space\_shuttle & 0.152 & 1.74 & 314 & 0.142 & 1.71 & 122 \\
    \midrule
    Mean           & 0.246 & 1.61 & 141 & 0.218 & 1.50 & 55 \\
    Median         & ---   & ---  & 101 & ---   & ---  & 53 \\
    \bottomrule
    \multicolumn{7}{@{}p{0.92\linewidth}@{}}{\footnotesize
      $^{*}$Trust-region fallback; the ARKit prior is scored
      (Section~\ref{sec:exp-failure}).}
  \end{tabular}
\end{table}

\section{Per-scene downstream rendering}
\label{sec:gsplat-perscene}

Table~\ref{tab:gsplat-perscene} breaks Table~\ref{tab:gsplat} down by
scene. Refinement improves PSNR over the ARKit prior on 13 of 18 scenes
for ours and 16 for the control; differences below ${\sim}0.4$\,dB sit
at the training-noise floor measured in Section~\ref{sec:exp-gsplat}.
Castle's ``Ours'' entry is the returned prior; the raw twisted solve is
analyzed in Section~\ref{sec:exp-failure}.

\begin{table}[t]
  \centering
  \caption{Per-scene held-out PSNR (dB) of 3D Gaussian splatting per
    pose source, identical training configuration throughout. Higher is
    better. Castle's ``Ours'' entry scores the returned prior
    (trust-region fallback).}
  \label{tab:gsplat-perscene}
  \small
  \setlength{\tabcolsep}{4pt}
  \begin{tabular}{lcccc}
    \toprule
    Scene & ARKit & Ours & COLMAP BA & GT \\
    \midrule
    aston          & 18.05 & 20.27 & 20.38 & 21.42 \\
    audi           & 14.55 & 14.16 & 14.23 & 14.13 \\
    beetles        & 14.09 & 14.79 & 14.90 & 14.75 \\
    big\_ben       & 13.87 & 16.18 & 16.49 & 17.02 \\
    boat           & 14.58 & 16.47 & 16.51 & 16.92 \\
    bridge         & 18.12 & 18.11 & 17.77 & 18.24 \\
    cabin          & 15.54 & 16.11 & 16.10 & 15.86 \\
    camera         & 17.59 & 17.50 & 17.66 & 18.67 \\
    castle         & 15.08 & 14.69 & 17.22 & 16.73 \\
    colosseum      & 16.23 & 18.14 & 17.74 & 18.04 \\
    convertible    & 16.95 & 18.35 & 18.48 & 19.71 \\
    ferrari        & 18.77 & 20.54 & 20.50 & 23.49 \\
    jeep           & 16.18 & 15.63 & 16.87 & 16.55 \\
    london\_bus    & 18.22 & 18.93 & 19.14 & 18.83 \\
    motorcycle     & 19.00 & 19.50 & 19.51 & 21.46 \\
    porsche        & 17.90 & 18.78 & 18.85 & 19.96 \\
    satellite      & 15.67 & 17.51 & 17.20 & 17.51 \\
    space\_shuttle & 17.44 & 18.95 & 19.27 & 19.93 \\
    \midrule
    Mean           & 16.55 & 17.48 & 17.71 & 18.29 \\
    \bottomrule
  \end{tabular}
\end{table}

\bibliographystyle{abbrvnat}
\bibliography{refs}

\end{document}